\documentclass[10pt,twocolumn,letterpaper]{article}

\usepackage[pagenumbers]{iccv} 

\definecolor{iccvblue}{rgb}{0.21,0.49,0.74}
\usepackage[pagebackref,breaklinks,colorlinks,allcolors=iccvblue]{hyperref}
\usepackage[accsupp]{axessibility}  

\def\paperID{2390} 
\def\confName{ICCV}
\def\confYear{2025}

\title{Balanced Image Stylization with Style Matching Score}

\author{Yuxin Jiang$^{1,2}$
\quad
Liming Jiang$^{3}$
\quad
Shuai Yang$^{4}$
\quad
Jia-Wei Liu$^{1}$
\quad
Ivor W. Tsang$^{2}$
\quad
Mike Zheng Shou$^{1\dagger}$\\\\
$^{1}$Show Lab, National University of Singapore
\quad
$^{2}$Agency for Science, Technology and Research (A*STAR)\\
\quad
$^{3}$Nanyang Technological University
\quad
$^{4}$Wangxuan Institute of Computer Technology, Peking University\\
}
\begin{document}

\twocolumn[{%
\renewcommand\twocolumn[1][]{#1}%
\maketitle
\begin{center}
    \vspace{-0.3cm}
\includegraphics[width=\linewidth]{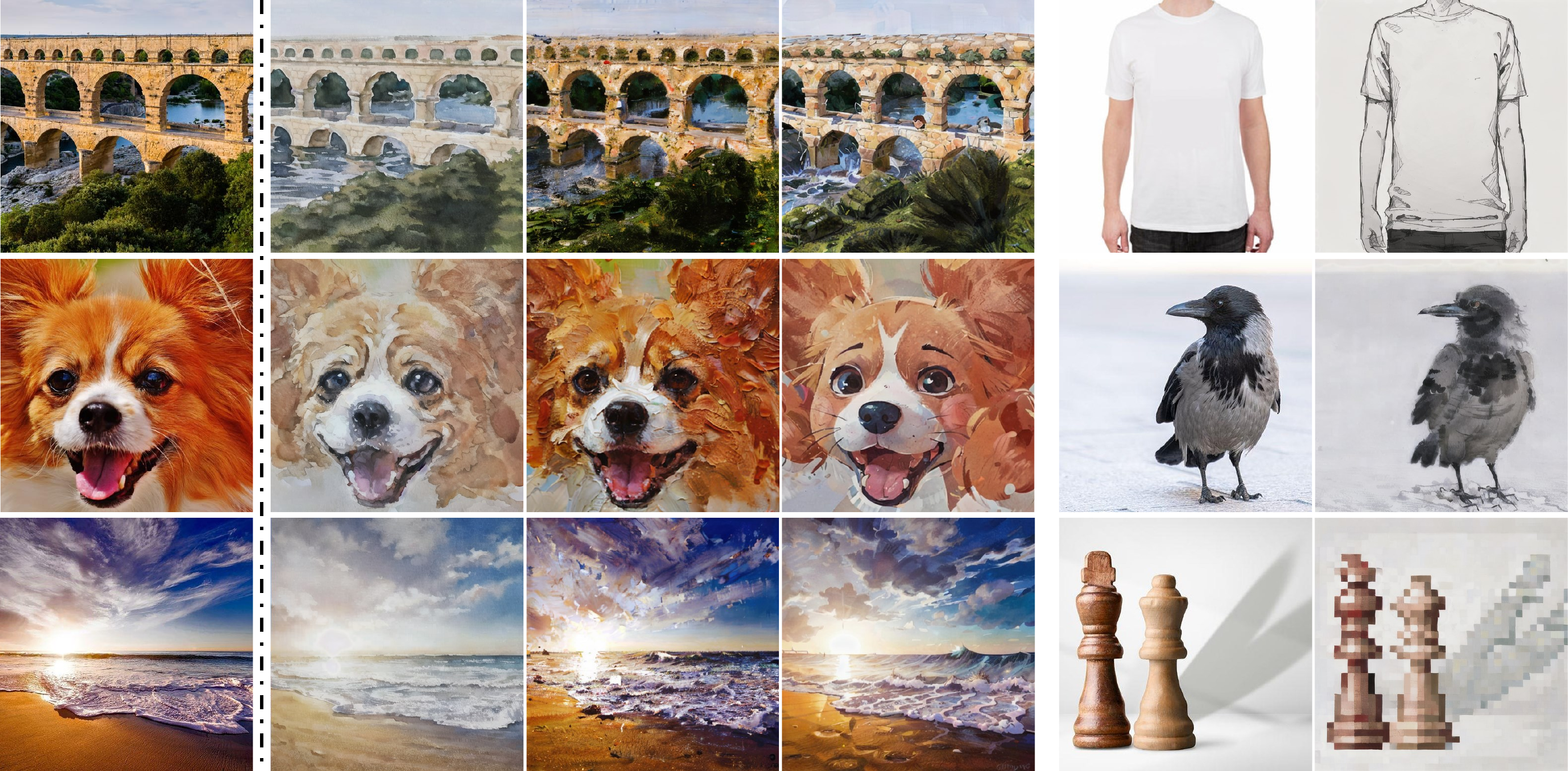}
\vspace{-5mm}
\captionof{figure}{\textbf{Examples of SMS applied to various artistic styles.} Stylized outputs (left to right, top to bottom) include watercolor, oil painting, kids' illustration, sketch, chinese ink painting, and pixel art style.}
\label{fig:teaser}
\end{center}%
}]

\maketitle

\begin{abstract}
We present Style Matching Score (SMS), a novel optimization method for image stylization with diffusion models. 
%
Balancing effective style transfer with content preservation is a long-standing challenge.
%
Unlike existing efforts, our method reframes image stylization as a style distribution matching problem. The target style distribution is estimated from off-the-shelf style-dependent LoRAs via carefully designed score functions. 
{\let\thefootnote\relax\footnotetext{$^\dagger$ Corresponding author.}}
%
To preserve content information adaptively, we propose Progressive Spectrum Regularization, which operates in the frequency domain to guide stylization progressively from low-frequency layouts to high-frequency details.
%
In addition, we devise a Semantic-Aware Gradient Refinement technique that leverages relevance maps derived from diffusion semantic priors to selectively stylize semantically important regions.
%
The proposed optimization formulation extends stylization from pixel space to parameter space, readily applicable to lightweight feedforward generators for efficient one-step stylization. 
%
SMS effectively balances style alignment and content preservation, outperforming state-of-the-art approaches, verified by extensive experiments.
{Code: \url{https://github.com/showlab/SMS}}.
\end{abstract}

\vspace{-5mm}    
\section{Introduction}
\label{sec:intro}

Image stylization transforms a content image to adopt a specific visual style, with applications in art design~\cite{gatys2016image}, virtual reality~\cite{hollein2022stylemesh}, and content creation~\cite{ruiz2024magic,everaert2023diffusion,saharia2022palette, song2025omniconsistency}. 
Early exemplar-based methods like Neural Style Transfer (NST)~\cite{gatys2016image, Johnson2016Perceptual} capture style using statistical correlations in CNN feature maps of a style image but are limited to single-image styles and are heavily content-influenced. 
Collection-based methods using GANs~\cite{zhu2017unpaired, chen2018cartoongan} model style as distributions over datasets for more realistic transfer but require large, well-curated datasets and face issues like mode collapse and training instability.

\begin{figure}[t]
\includegraphics[width=\linewidth]{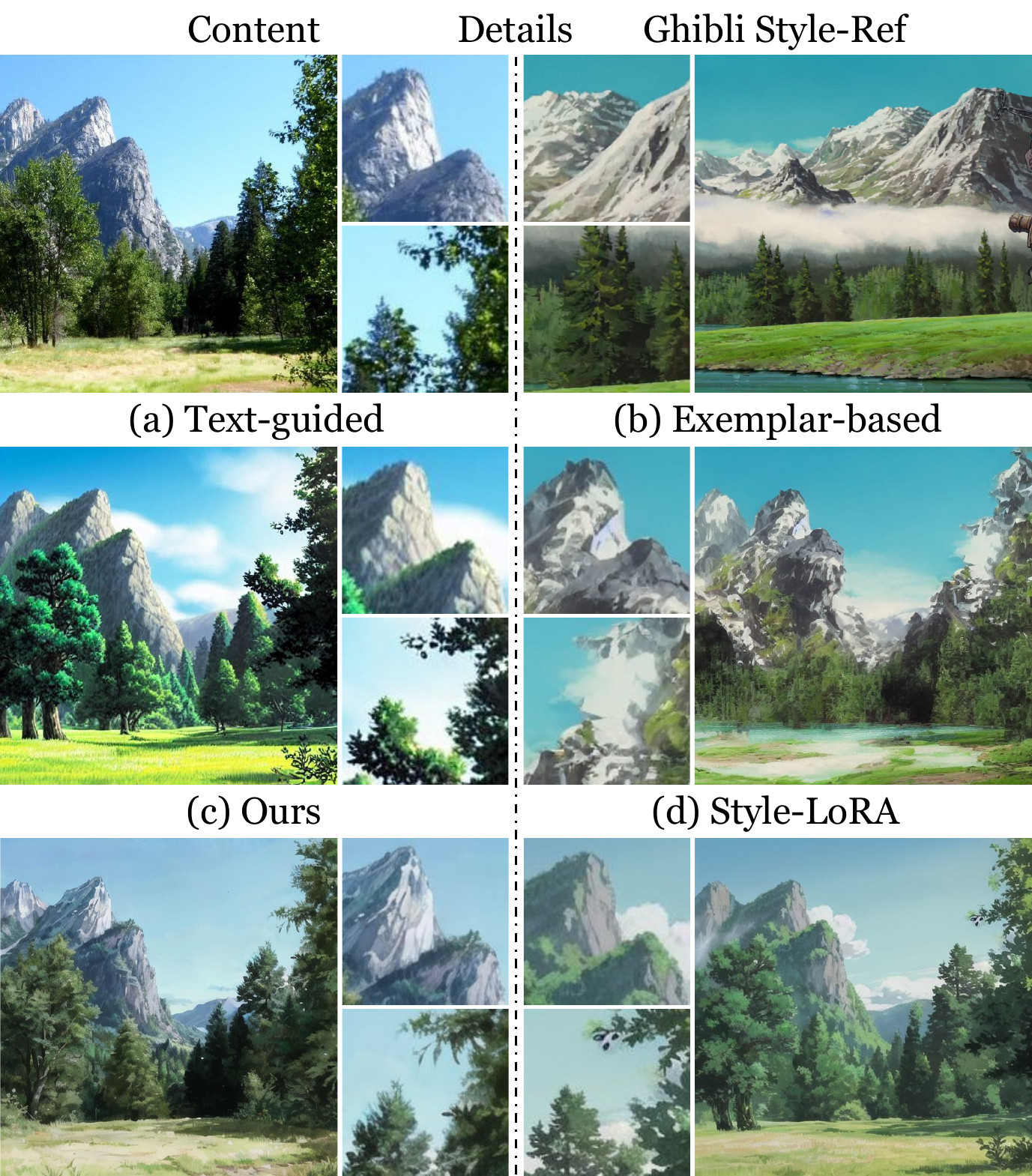}
\vspace{-4mm}
\caption{\textbf{Comparison of different style representations. }
(a) Fails to mimic the Ghibli style. (b) Introduces style elements but lack spatial coherence, simply overlaying textures from the reference style. (d) Captures Ghibli-like textures but distorts content and color due to limitations in ControlNet~\cite{zhang2023adding} conditioning. (c) Achieves a harmonious balance of content preservation and style adaptation with fine local stylized details.
}
\vspace{-5mm}
\label{fig:styletexture}
\end{figure}

The emergence of multimodal models like CLIP~\cite{radford2021learning}, and diffusion models (DMs), such as Stable Diffusion (SD)~\cite{rombach2022high}, has ushered in a new era for image style transfer.
Recent approaches can be categorized into zero-shot text-guided, one-shot exemplar-based, and collection-based fine-tuning methods. 
Zero-shot text-guided methods~\cite{yang2023zero, he2024freestyle} use text prompts for style transfer but often struggle to accurately capture complex styles that are challenging to describe with words (see Figure~\ref{fig:styletexture}(a)), as ``one image is worth a thousand words''. 
One-shot exemplar-based methods~\cite{zhang2023inversion, Chung_2024_CVPR, wang2024instantstyle}, similar to traditional NST, struggle to model an artist's overall style and often overemphasize the exemplar's style, leading to content distortion (see Figure~\ref{fig:styletexture}(b)).
Collection-based fine-tuning methods, such as those using LoRA~\cite{hu2022lora}, adapt DMs to new styles with minimal retraining, effectively capturing intricate patterns and an artist's overall style. 
However, these methods often overemphasize style at the expense of content preservation, struggling to balance style and content. 
When conditioned on specific content images, they typically use ControlNet~\cite{zhang2023adding} with edge maps as conditions or DDIM inversion~\cite{songdenoising} and generate stylized images from noises, which may still fail to preserve sufficient content details (see Figure~\ref{fig:styletexture}(d)).

To address these challenges,
 we propose the \textit{Style Matching Score (SMS)}, a novel optimization method that reframes style transfer as style distribution matching. 
Specifically, we introduce: \textit{1) Style Matching Objective} for transferring style; \textit{2) Progressive Spectrum Regularization} for preserving content; and \textit{3) Semantic-Aware Gradient Refinement} for balancing style and content with diffusion semantic priors.
Our key idea is to match the distribution of the output images with the target style distribution by leveraging powerful diffusion priors.
We minimize the Kullback–Leibler (KL) divergence between the distribution of the stylized images and the target style distribution, estimated by score functions of a style LoRA-integrated pretrained DM. 
By interpreting the denoised gradient directions that make the image more stylized, we effectively align generated images with the target style.
At a high level, our method shares the motivation of score distillation methods~\cite{wang2023prolificdreamer, yin2024onestep}. 
We differ by focusing on distilling style information into the real image domain, where preserving the identity of the source content is a key challenge.

To achieve this, we incorporate an explicit progressive identity regularization in frequency domain.
%
Observing that stylistic differences largely reside in high-frequency components, we guide stylization progressively from low-level structures to high-frequency details, ensuring structural integrity and harmonious transitions. 
This is achieved by preserving more low-frequency components during high-noise steps and allowing finer stylized details at lower-noise steps. Combined with an adaptive narrowing sampling strategy, our method minimizes content disruption while retaining sufficient style details.
%
%
To further balance style and content, we introduce a semantic-aware gradient refinement that leverages the relevance map~\cite{mirzaei2023watchyoursteps} derived from DM's semantic priors.
Our key insight is that each pixel requires different degrees of stylization.
By uniquely using the map as an element-wise weighting mechanism, we allow gradients in more semantically-important regions to emphasize stylistic transformation while attenuating gradients in less critical areas to preserve content integrity. 
%
%

Thanks to our optimization-based formulation, SMS extends stylization from the pixel space to the parameter space, enabling broader and more flexible style transfer capabilities.
For instance, SMS is readily applicable to a lightweight feed-forward generator, allowing for efficient one-step generation, potentially useful for more applications beyond.
%
In summary, our contributions are fourfold: 
\begin{itemize}
\item We introduce a novel SMS optimization method to leverage diffusion priors to match style distributions for semantic-aware and content-preserving style transfer.
\item We develop a progressive spectrum regularization term  and adaptive narrowing sampling strategy guided by diffusion characteristics, to preserve content and achieve fine-detail stylization.  
\item We propose an semantic-aware gradient refinement to balance style and content by selectively stylizing semantically-important regions. 
\item We demonstrate SMS is readily applicable to distill style into lightweight feed-forward networks, extending applicability beyond single-image stylization.
\end{itemize}

\section{Related Work}
\label{sec:relatedwork}

\noindent {\bf Image Style Transfer} aims to synthesize images in artistic styles and has significant practical value.
%
GAN-based methods~\cite{zhu2017unpaired, chen2018cartoongan, jiang2023scenimefy} enable realistic stylization by learning style distributions from datasets, which however requires extensive data and are challenging to train.
%
Recent work leverages pretrained diffusion models (DMs) for style transfer, categorized as follows. 
\textit{1) Prompts-Guide Stylization}, represented by FreeStyle~\cite{he2024freestyle}, offers flexibility but struggles with ambiguous or imprecise style control due to the abstract nature of text guidance.
%
\textit{2) Exemplar-Based Stylization} typically manipulates self-attention or cross-attention layers to transfer style from a reference image.
InST~\cite{zhang2023inversion} maps style images to textual embeddings for conditions. 
StyleID~\cite{Chung_2024_CVPR} maintains content integrity through query preservation and AdaIN~\cite{huang2017arbitrary}.
InstantStyle-Plus~\cite{wang2024instantstyle} injects style-specific features into intermediate model layers.
Single-style exemplars, however, lead to over-stylization, obscuring semantic content.
\textit{3) Tuning-Based Stylization} fine-tunes DMs on target style distributions, which produces high-quality results but is computationally costly.
Style-dependent LoRA adapt models to specific styles with less data and capture nuanced style distributions~\cite{civitai}.
However, they typically use ControlNet~\cite{zhang2023adding} with basic edge maps from the source image for image-to-image stylization, which may lose details.
To overcome these limitations, we propose a distribution-level style matching loss, fully leveraging diffusion priors for semantic-aware, identity-preserving and applicable stylization.

\noindent {\bf Score Distillation}, represented by Score Distillation Sampling (SDS)~\cite{pooledreamfusion} and Variational Score Distillation (VSD)~\cite{wang2023prolificdreamer}, is initially introduced in the context of text-to-3D synthesis by distilling the 2D generative prior of text-to-image DMs. Later, researchers have applied this idea to diffusion distillation, such as  Distribution Matching Distillation (DMD)~\cite{yin2024onestep} and text-driven image editing~\cite{hertz2023delta,koo2024posterior,nam2024contrastive}.
However, these SDS-based image editing methods mainly overemphasize the style and fail to preserve the content for image stylization. To solve this issue, 
Delta Denoising Score (DDS)~\cite{hertz2023delta} reduces noisy gradient directions in SDS to better maintain the input image details.  
Building on DDS, Posterior Distillation Sampling (PDS)~\cite{koo2024posterior} introduces a stochastic latent matching loss to add an explicit spatial identity preservation regularization term,
while contrastive Denoising Score (CDS)~\cite{nam2024contrastive} adds a contrastive loss to maintain identity.
Despite efforts, DDS suffers from content preservation due to increasing mismatch in source-target distribution~\cite{mcallister2024rethinking}, as reduced noisy denoising directions are not calculated based on the current optimized image
%
while VSD effectively minimizes the mismatch error by test-time finetuning a copy of the DM with LoRA on the current set (\textit{i.e.}, the fake distribution).
Our method shares the high-level objectives as VSD and DMD but prioritizes identity preservation for stylization. 
We introduce an semantic-aware gradient refinement and spectrum regularization, specifically tailored for style transfer, ensuring precise stylization.

\section{Style Matching Score}
\label{sec:SMS}

\subsection{Problem Modeling}

Given a source image ${x}^{\text{src}} \in p_{\text{real}}$, a text description $y_{\text{src}}$, and a fixed target style distribution $p_{\text{style}}$, our goal is to synthesize a stylized image ${x}^{\text{tgt}}$ such that:
\begin{enumerate}
    \item \textit{Style Alignment:} aligns with the style distribution  $p_{\text{style}}$.
    \item \textit{Content Preservation:} retains the identity of ${x}^{\text{src}}$.
\end{enumerate}
To model this problem flexibly, we parameterize the image generation process using $\theta$. Specifically, we define a parametric generator $G_\theta$ such that ${x}^{\text{tgt}} = G_\theta({x}^{\text{src}})$. Here, $\theta$ can represent either the image itself (in test-time optimization) or the parameters of an image-to-image generator network.

To achieve style alignment, we aim to minimize the KL divergence between the generated distribution $p_{G_\theta}$ and target style distribution $p_{\text{style}}$:
\begin{equation} 
\label{eq:kl}
D_{\text{KL}}(p_{G_\theta}|| p_{\text{style}}) = \int p_{G_\theta}(x^{\text{tgt}}) \log \frac{p_{G_\theta}(x^{\text{tgt}})}{p_{\text{style}}(x^{\text{tgt}})}dx.
\end{equation}

\subsection{Style Matching Objective}
\label{sec:stylematching}
Following DMD~\cite{yin2024onestep}, we estimate the distributions in Eq.~(\ref{eq:kl}) using score functions approximated by diffusion models.
Specifically, we employ two noise prediction models $\epsilon_\text{style}$ (fixed) and $\epsilon_{\text{fake}}^\phi$ (dynamically learned with $L_{\text{denoise}}^\phi$~\cite{yin2024onestep}), which approximate the score functions of $p_{\text{style}}$ and $p_{G_\theta}$, respectively.
%
Then, $\nabla_\theta D_{\text{KL}}$, the gradient of $D_{\text{KL}}$ with respect to $\theta$, is approximated as:
\begin{equation} 
\label{eq:dmd} 
\underset{t, \epsilon}{\mathbb{E}}  \left[ w_t \left( \epsilon_{\text{style}}(z_{t}^\text{tgt}; y_{\text{src}}, t) - \epsilon_\text{fake}^\phi(z_{t}^\text{tgt}; y_{\text{src}}, t) \right) \frac{\partial G_\theta}{\partial \theta} \right], 
\end{equation}
where $y_{\text{src}}$ is the text prompt describing the content of the source image ${x}^{\text{src}}$, $z_{t}^\text{tgt} = \sqrt{\bar{\alpha}_t} z_{0}^\text{tgt} + \sqrt{1 - \bar{\alpha}_t} \epsilon$, $\epsilon \sim \mathcal{N}(0, \mathbf{I})$, and $z_{0}^\text{tgt} = \varepsilon(G_\theta(x^{\text{src}}))$. The function $\varepsilon(\cdot)$ is the VAE encoder of the SD, and $w_t$ is a timestep-dependent scalar weight. 
Intuitively, the denoising direction $\epsilon_\text{style}$ moves $z_{0}^\text{tgt}$ toward the modes of $p_{\text{style}}$, while $-\epsilon_{\text{fake}}^\phi$ spreads them apart, steering the image towards a desired stylized direction. For a detailed derivation, please refer to~\cite{yin2024onestep} and our appendix. 

Our approach differs from DMD~\cite{yin2024onestep}, which employs a general pretrained DM $\epsilon_{\text{real}}$ without style specialization to represent the target distribution. 
In contrast, we tailor the DM to the target style by integrating a style-specific LoRA module into $\epsilon_{\text{real}}$, resulting in $\epsilon_{\text{style}}$. 
This integration harnesses the superior style modeling capabilities of the style-LoRA, bridging the gap between general DM and specific style needs.
Our method captures the nuances of the target style more accurately, enhancing stylization fidelity and better aligning the generated images with the desired artistic characteristics.

%
%
\begin{figure}[t]
    \centering
    \includegraphics[width=\linewidth]{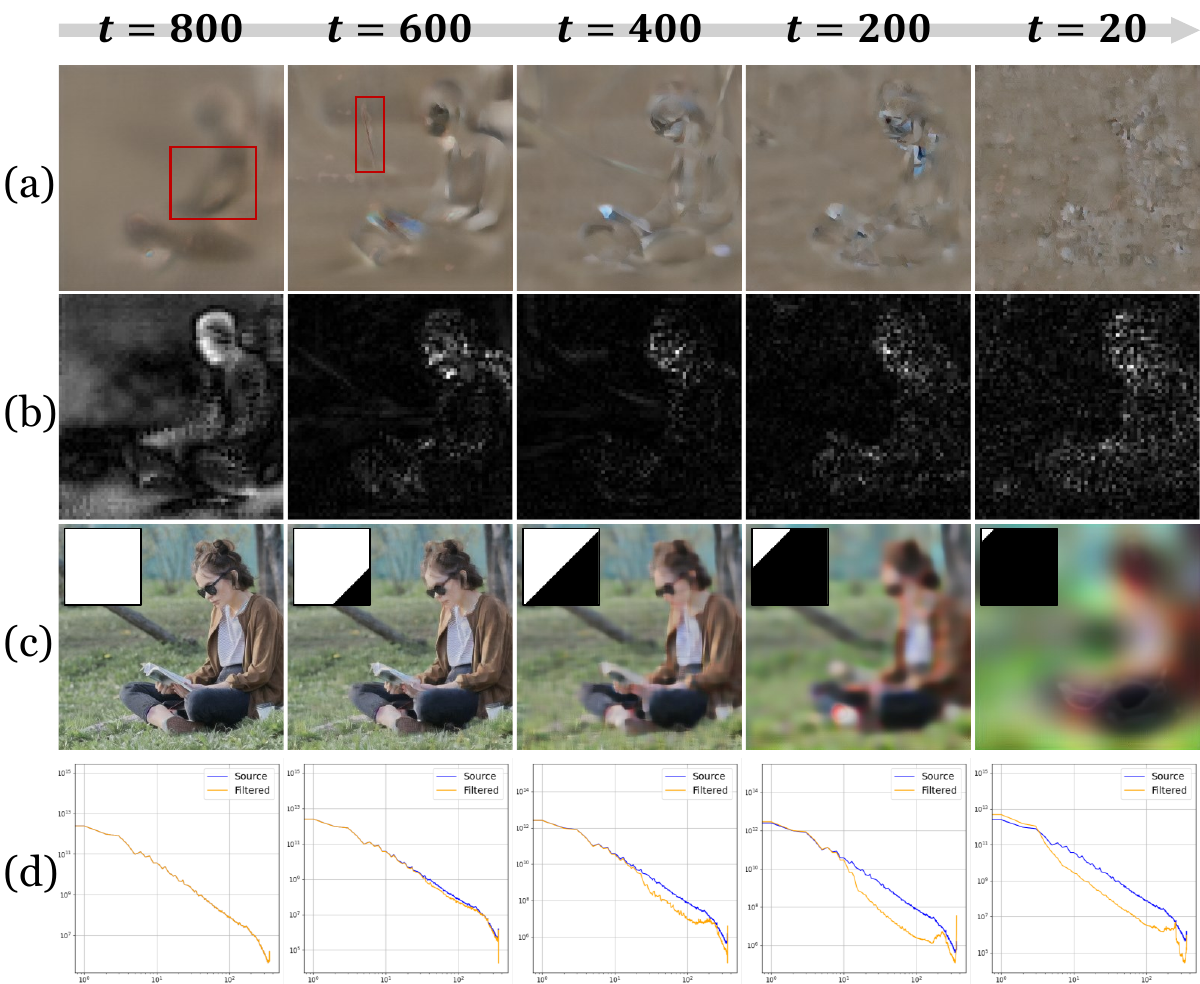}
    \vspace{-5.5mm}
    \caption{{\bf Visualization across different timesteps.}
    (a) Style matching gradient $\nabla_\theta D_{\text{KL}}$; (b) Relevance Coefficient $\mathcal{R}(z_t^{\text{src}}, t)$; (c) Regulated frequency component in the spatial domain $\text{IDCT}(\mathcal{F}_{\text{low}})$ with the corresponding low-pass filter mask $\text{LPF}(\cdot)$;
    (d) Radial average power spectrum comparing the source image's frequency distribution (blue) with the regulated component (orange).} 
    \vspace{-5mm}
    \label{fig:rel_map}
\end{figure} 
\begin{figure*}[!ht]
    \centering
    \includegraphics[width=\linewidth]{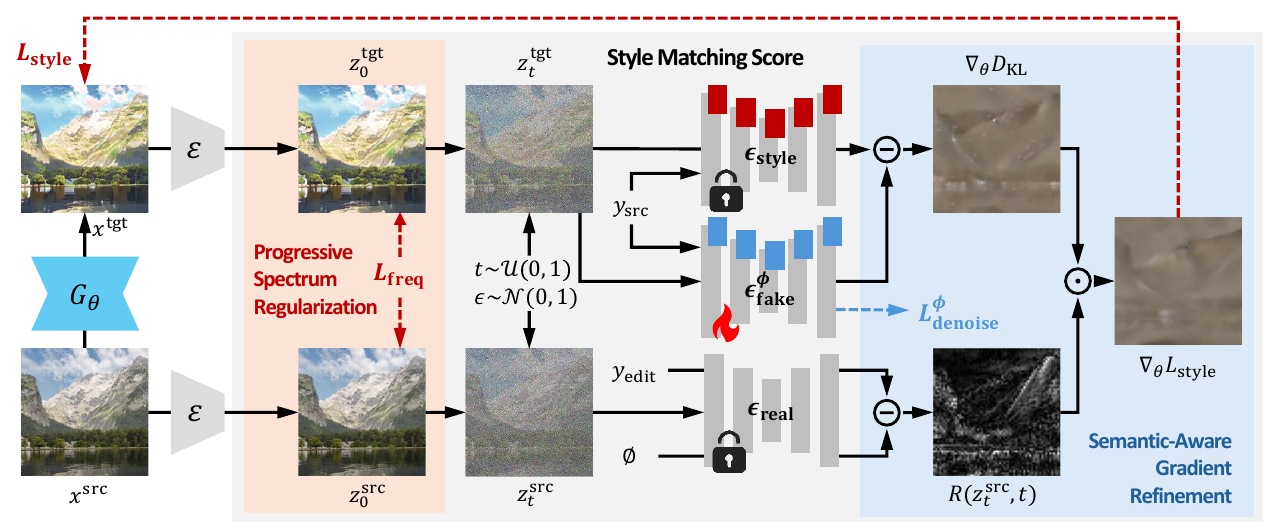}
    \caption{\textbf{Overview of SMS for Feed-Forward Stylization.} We train a one-step generator $G_\theta$ to map source images $x^\text{src}$ into stylized outputs.
    To transfer style, we compute the style matching gradient $\nabla_\theta D_{\text{KL}}$ by injecting random noise into the generated image and passing it through two diffusion models: one integrated with Style-LoRA (representing the target style distribution) and one continually trained on the generated images (representing the current ``fake'' distribution). 
    The difference between their denoising scores provides a gradient direction to make the image more stylized and less fakeness.
    A relevance map, derived from the disagreement between noise predictions with and without stylized instructions, acts as an element-wise correction to selectively stylize semantically important regions, ensuring fidelity and coherence. 
    Finally, progressive spectrum regularization is applied to preserve content within the latent space.}
    \vspace{-5mm}
    \label{fig:framework}
\end{figure*}

\subsection{Progressive Spectrum Regularization}

Unlike DMD, which maps from noise to images $z_{\theta, 0} = \varepsilon(G_\theta(\eta))$ with $\eta \sim \mathcal{N}(0, \mathbf{I})$, our style matching objective focuses on transforming the real image distribution to the style distribution, required preserving content correspondence.
However, because the text prompt $y_{\text{src}}$ allows for a probabilistic distribution of all possible outputs, the raw gradient $\nabla_{\theta} D_{KL}$ can be noisy, introducing unrelated changes (see Figure~\ref{fig:rel_map}(a)).

To mitigate the loss of fidelity, we introduce explicit content regularization in frequency domain. The analysis (see our appendix) indicates that the primary differences between real and stylized images lie in the high-frequency components.
Previous methods that directly regulate in the spatial domain either fail to preserve identity or apply too strict regularization, limiting effective style transfer.
In contrast, our method progressively guide stylization from low-level structures to high-frequency details using a timestep-aware spectrum.
Specifically, we preserve more low-frequency components during high-noise steps (large $t$) for structural integrity and allow greater flexibility in high-frequency components during low-noise steps (small $t$) to enable detailed stylization. 
The progressive spectrum regularization term is defined as:
\begin{equation} 
\label{eq:dct} 
L_{\text{freq}} = \Big|\Big| \mathcal{F}_{\text{low}}(z_{0}^\text{tgt}, t) - \mathcal{F}_{\text{low}}(z_{0}^\text{src}, t) \Big|\Big|_2^2,
\end{equation}
where $\mathcal{F}_{\text{low}}(z, t) = \text{LPF}(\text{DCT}(z), \text{thld}(t))$, where LPF denotes a low-pass filter. Specifically, $\text{LPF}(\cdot, \text{thld}(t))$ applies a low-pass filter to the Discrete Cosine Transform (DCT) of $z$ using a cutoff frequency threshold determined by $t$, denoted as $\text{thld}(t)$. As shown in Figure~\ref{fig:rel_map}(c), $\text{thld}(t)$ decreases with $t$, resulting in progressively looser regularization for step-aware content preservation.

\subsection{Semantic-Aware Gradient Refinement}
To further balance style and content, we introduce a semantic-aware gradient refinement mechanism.
Recognizing that not every pixel of an image require the same degree of stylization; for example, foreground subjects might need stronger stylization, whereas background elements benefit from minimal alteration to preserve identity, we leverage the DM's semantic priors to computer a relevance map~\cite{mirzaei2023watchyoursteps} to guide gradient modulation.
Specifically, we define the relevance coefficient $\mathcal{R}(z_t^{\text{src}}, t)$ at timestep $t$ as:
\begin{equation}
\label{eq:rel-coefficeint} 
\mathcal{R}(z_t^{\text{src}}, t) = \text{Norm} \big( \big| \epsilon_{\text{real}}(z_t^{\text{src}}; y_{\text{edit}}, t) - \epsilon_{\text{real}}(z_t^{\text{src}}; y_{\emptyset}, t) \big| \big),
\end{equation}
where $z_t^{\text{src}} = \sqrt{\bar{\alpha}_t} z_0^{\text{src}} + \sqrt{1 - \bar{\alpha}_t} \epsilon$, $\epsilon \sim \mathcal{N}(0, \mathbf{I})$, and $ z_0^{\text{src}} = \varepsilon(x^{\text{src}})$. 
$y_{\text{edit}}$ is the editing instructions specifying the desired style, and $y_\emptyset$ is the empty condition. The absolute difference $|\cdot|$  between these predictions highlights regions where the model anticipates changes due to the style instructions. The function $\text{Norm}(\cdot)$ applies min-max normalization to scale the relevance map to range $[0, 1]$. 

This relevance map effectively captures the semantic importance of each pixel with respect to the style transformation. 
By employing it as an element-wise weighting mechanism, we modulate the gradient to emphasize stylistic changes in  semantically significant regions while attenuating changes in less critical areas.
The refined style matching loss is:
\vspace{-2mm}
\begin{equation} 
\label{eq:style} 
\begin{split}
L_\text{style} = \underset{t, \epsilon}{\mathbb{E}} \Big[ \Big|\Big| \mathcal{R}(z_t^{\text{src}}, t) \odot w_t 
& \Big( \epsilon_{\text{style}}(z_t^\text{tgt}; y_\text{src}, t) \\
& - \epsilon_\text{fake}^\phi(z_t^\text{tgt}; y_\text{src}, t) \Big) \Big|\Big|_2^2 \Big],
\end{split}
\end{equation}
where $\odot$ denotes element-wise multiplication. 
Moreover, the relevance coefficient is adaptive, timestep-depedent, it aligns with the progression of the diffusion process, focusing gradient modulation on appropriate regions at different step (see Figure~\ref{fig:rel_map}(b)). This dynamic adjustment allows for controlled and semantic-aware coherent style transfer across optimization.

\subsection{Overall Training Objective} 
Combining the progressive spectrum regularization and semantic-aware gradient refinement, our overall training objective is:
\vspace{-4mm}
\begin{equation}
\label{eq:sms} 
L_{\text{SMS}} = L_{\text{style}} + \lambda \cdot L_{\text{freq}},
\end{equation}
where $\lambda$ is the loss weight. 
By minimizing $L_{\text{SMS}}$, we guide $G_\theta$ to produce images that align with the target style in semantically important regions while retaining the identity of the source image.

%
\noindent \textbf{Adaptive Narrowing Sampling Strategy.}
Uniformly random sampling of timesteps $t$ results in inconsistent regularization strength across iterations, leading to blurred images and ineffective stylization.
Conversely, naive linear annealing of $t$ cause excessive high-frequency editing in later stages, producing deviations from the original content due to looser content restrictions.
To address this, we propose an adaptive narrowing sampling strategy that gradually reduces the upper bound of the timestep sampling range over iterations while preserving randomness, effectively integrating with our progressive spectrum regularization.
Specifically, we sample $t$ as:
\vspace{-2mm}
\begin{equation} 
\label{eq:timestep} 
t \sim \mathcal{U}(t_{\min}, t_{\text{upper}}), 
\quad t_{\text{upper}} = \left( 1 - \frac{\text{iter}_\text{cur}}{\text{iter}_\text{total}} \right) \cdot t_{\max},
\end{equation}
where $\text{iter}_\text{cur}$ is the current iteration and $\text{iter}_\text{total}$ is the total number of iterations.
%


\begin{figure*}[!ht]
    \centering
    \vspace{-4mm}
    \includegraphics[width=0.96\linewidth]{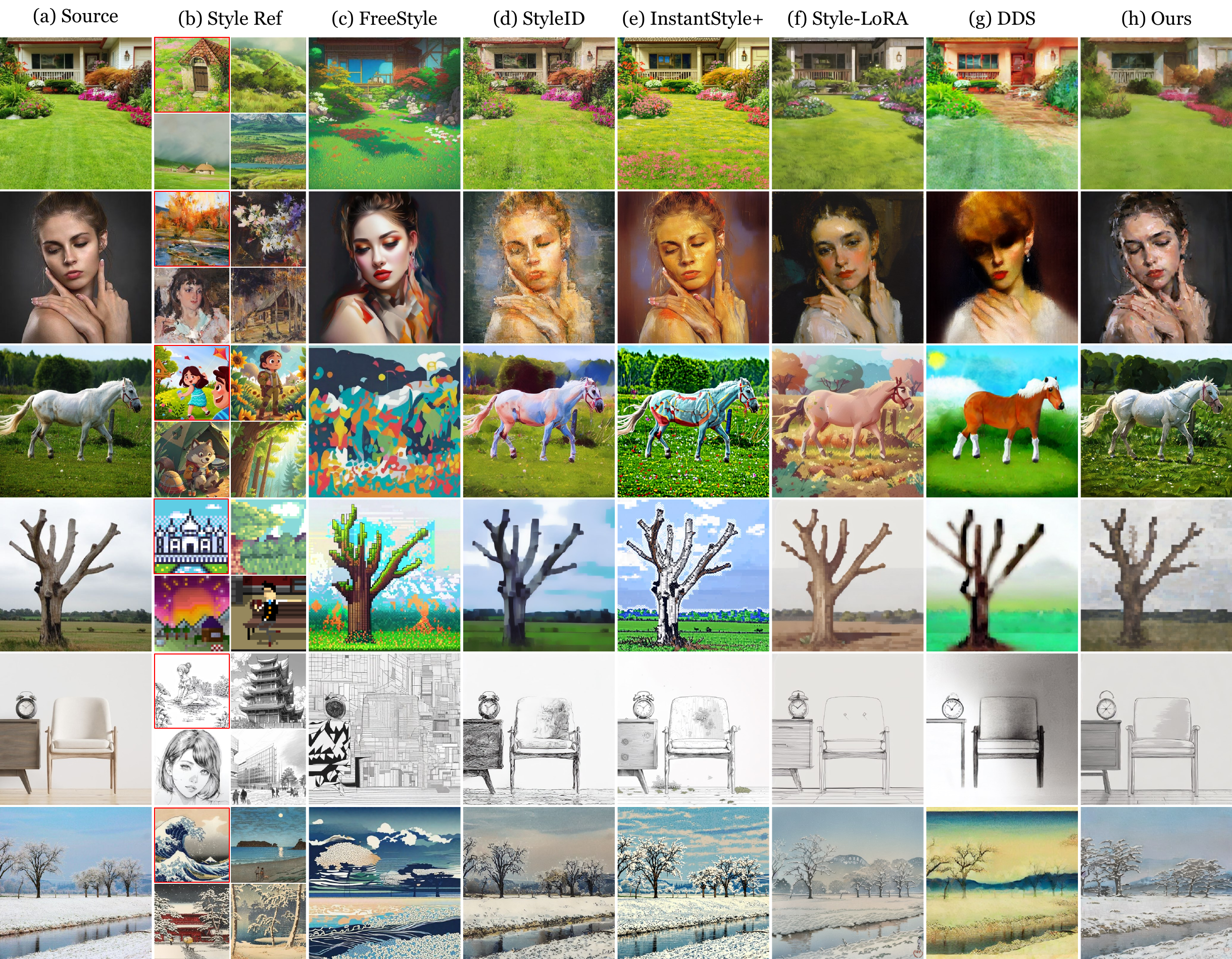}
    \vspace{-2mm}
    \caption{{\bf Qualitative comparison.} We compare SMS (Ours) with five representative methods. Our approach achieves superior semantic consistency and style texture fidelity, striking the best balance between style alignment and content preservation compared to state-of-the-art baselines.  The style references in the red boxes are used as the exemplars for exemplar-guided methods. Please zoom in for details}
    \vspace{-1mm}
    \label{fig:mainresults}    
\end{figure*}

\newcommand{\tablefirst}[0]{\cellcolor{myred}}
\newcommand{\tablesecond}[0]{\cellcolor{myorange}}
\newcommand{\tablethird}[0]{\cellcolor{myyellow}}

\vspace{-2mm}
\begin{table*}[!ht]
\centering
\caption{\normalsize Quantitative comparison with diffusion-based baselines using different style representations. The best results are highlighted in \textbf{bold}, and the second-best are \underline{underlined}.}
\vspace{-2mm}
\label{tbl:combined}
\resizebox{\textwidth}{!}{%
\begin{tabular}{c|c|cccccc|c|cccccccc}
\hline
\multirow{2}{*}{Metric} & \multicolumn{7}{c|}{Ghibli Style} & \multicolumn{7}{c}{Oil Painting Style} \\
\cline{2-15}
 & Real 
 & \makecell{FreeStyle\\~\cite{he2024freestyle}} 
 & \makecell{StyleID\\~\cite{Chung_2024_CVPR}} 
 & \makecell{InstantStyle\\-Plus~\cite{wang2024instantstyle}} 
 & \makecell{Style\\-LoRA}
 & \makecell{DDS\\~\cite{hertz2023delta}} 
 & Ours 
 & Real 
 & \makecell{FreeStyle\\~\cite{he2024freestyle}} 
 & \makecell{StyleID\\~\cite{Chung_2024_CVPR}} 
 & \makecell{InstantStyle\\-Plus~\cite{wang2024instantstyle}} 
 & \makecell{Style\\-LoRA}
 & \makecell{DDS\\~\cite{hertz2023delta}} 
 & Ours \\
\hline
LPIPS $\downarrow$ & 0.000 & 0.690 & 0.608 & 0.538 & \underline{0.438} & 0.513 & \textbf{0.326} & 0.000 & 0.686 & 0.626 & \underline{0.524} & 0.588 & 0.569 & \textbf{0.431} \\
CFSD $\downarrow$ & 0.000 & 0.786 & 0.139 & 0.337 & \underline{0.113} & 0.129 & \textbf{0.102} & 0.000 & 1.083 & 0.156 & 0.138 & \underline{0.129} & 0.130 & \textbf{0.128} \\
FID $\downarrow$ & 15.138 & \underline{12.361} & 19.007 & 14.949 & \textbf{12.267} & 15.233 & 13.089 & 30.124 & 29.974 & \textbf{16.252} & 24.858 & 21.861 & 23.133 & \underline{17.905} \\
ArtFID $\downarrow$ & 16.138 & 22.582 & 32.169 & 24.532 & \underline{19.077} & 24.554 & \textbf{18.686} & 31.124 & 52.218 & \underline{28.046} & 39.402 & 36.292 & 37.864 & \textbf{27.055} \\
PickScore $\uparrow$ & 1.000 & 0.683 & 0.405 & 1.019 & \textbf{2.067} & 0.537 & \underline{1.487} & 1.000 & 0.247 & 0.803 & 1.350 & \textbf{2.862} & 0.782 & \underline{1.842} \\
HPSv2 $\uparrow$ & 0.272 & 0.201 & 0.210 & 0.257 & \textbf{0.264} & 0.227 & \underline{0.259} & 0.260 & 0.164 & 0.224 & 0.258 & \textbf{0.278} & 0.231 & \underline{0.260} \\
\hline
\end{tabular}
}
\vspace{-4mm}
\end{table*}


\section{Feed-Forward Stylization with SMS}
\label{sec:FFN}

As an applications of SMS, we present a detailed data-free stylizaton pipeline using a lightweight feed-forward network $G_\theta$ (see~\Cref{fig:framework}).
Given a set of real images $\{ x_i \}_{i=1}^N \in p_\text{real}$ and a target style domain $p_\text{style}$ modeled by a pretrained $\epsilon_\text{style}$, our goal is to learn a mapping $G_{\theta} : p_\text{text} \rightarrow p_\text{style}$ that transforms input images into the target style domain.
\noindent \textbf{Reconstruction Warmup.} 
To ensure that $G_\theta$ maintains content fidelity and accelerates convergence, we initialize $G_\theta$ by training it to reconstruct the input images. 
This warmup phase aligns the initial generated distribution $p_{G_\theta}$ with the real image distribution $p_\text{real}$, providing a solid foundation for subsequent training. 
The reconstruction loss is defined as:
The reconstruction loss is defined as: 
\begin{equation} 
\label{eq:rec}
\mathcal{L}_{\text{rec}} = \underset{x \sim p_\text{real}}{\mathbb{E}} \Big[ \big| \big| G_{\theta}(x) - x \big| \big|_1 \Big], 
\end{equation}
where $|| \cdot ||_1$ denotes the $L1$ norm.

\noindent \textbf{Per-Batch Variable Timesteps.}
After warmup, we train $G_\theta$ using the SMS loss (see Eq.~(\ref{eq:sms})).  
Unlike direct pixel updates in single-image optimization, updates to $G_\theta$ affect the generated distribution indirectly through changes in the network parameters.
Additionally, network training requires generalization to diverse inputs and robustness to variations.
To address these, we introduce per-batch variable timesteps.
For each training batch, we use the same image $x$ repeated $B$ times (where $B$ is the batch size).
We add different levels of noise corresponding to different timesteps $\{t_i\}_{i=1}^B$ to each instance of the image within the batch:
\begin{equation} 
\label{eq} 
z_{t_i} = \sqrt{\bar{\alpha}{t_i}} z_0 + \sqrt{1 - \bar{\alpha}{t_i}} \epsilon_i, 
\end{equation}
where $z_0 = \mathcal{E}(G_\theta(x))$. $\epsilon_i \sim \mathcal{N}(0, \mathcal{I})$ are independent noise samples, and $t_i$ are independently sampled timesteps for each instance in the batch.
By computing gradients over multiple noise levels for the same image, we effectively average the optimization directions, leading to more stable and consistent updates to $G_\theta$.
We find that our proposed strategy of gradually reducing the sampling range of timestep also works well in this case.

\section{Experiments}
\label{sec:exp}

%
%

\subsection{Single-Image Stylization}
\label{sec:single}
\noindent \textbf{Baselines.}
We compare our method against five state-of-the-art diffusion-based style transfer methods categorized by their style representations:
1) Zero-shot text-driven: FreeStyle~\cite{he2024freestyle}, DDS~\cite{hertz2023delta}; 2) One-shot Exemplar-guided: StyleID~\cite{Chung_2024_CVPR}, InstantStyle-Plus~\cite{wang2024instantstyle}; 3)  Collection-based: Style-specific LoRA generation with ControlNet integrated for spatial preservation (referred to as Style-LoRA).

\noindent \textbf{Datasets.}
We use the PIE-benchmark~\cite{ju2023direct}, containing $280$ diverse images covering animals, people, indoor and outdoor scenes, each paired with textual descriptions.

\noindent \textbf{Implementation Details.}
We conduct all experiments using Stable Diffusion $1.5$. The relevance map's editing instruction $y_{edit}$ is set as ``Turn it into \textit{\{target style\}}'', with a total of 500 training iterations. The DCT threshold frequency $\text{thld}(t)$ is defined as $t/500$. Sampling $t$ ranges from $[20, 500]$ following our adaptive narrowing sampling strategy.
Baseline methods are implemented using their official code and default settings.
More implementation details are provided in the appendix.

%
%

\noindent \textbf{Evaluation Metrics.}
We quantitatively evaluate our method on two representative artistic styles: \textit{``Ghibli style''}, a renowned Japanese anime style known for its intricate semantic-aware style textures, and \textit{``Oil Painting style``}, characterized by its coarser, larger brush strokes. 
%
We assess style transfer performance across three aspects: content preservation, style fidelity, and overall image quality. 
Content preservation is measured using LPIPS~\cite{zhang2018unreasonable} and CFSD~\cite{Chung_2024_CVPR}, evaluating structural similarity between stylized and corresponding source images. 
Style fidelity is quantified via FID~\cite{heusel2017gans} against authentic Ghibli images from original films. 
Overall perceptual quality is evaluated by ArtFID~\cite{wright2022artfid}, defined as $(\text{LPIPS} + 1) \cdot (\text{FID} + 1)$, alongside aesthetic metrics such as PickScore~\cite{kirstain2023pick} and Human Preference Score (HPS)~\cite{wu2023human}.
%
Additionally, we conduct a user study to further validate the balance and effectiveness of our method. Participants are asked to choose the best result in terms of three criteria: evident style (Style), content preservation (Content), and overall translation quality (Balance). Higher scores indicate better image quality. 

\noindent \textbf{Qualitative Comparison.}
As illustrated in Figure~\ref{fig:mainresults}, our approach excels at balancing style and content.
FreeStyle over-shifts the style, losing identity preservation.
Exampler-based methods maintain content integrity through DDIM inversion but lack sufficient style infusion, appearing as texture overlays and noticable color shifts.
Style-LoRA effectively captures stylistic textures but struggle with content and color preservation due to limitations in ControlNet conditioning. 
DDS faces challenges in transferring style using text prompts alone, leading to content deviations and over-saturated colors due to the absence of explicit identity regularization. 
SMS, in contrast, presents delicate, semantic-aware style features while retaining strong identity preservation.
The qualitative results indicate the effectiveness of our method, surpassing the state-of-the-art baselines in perceptual quality for single-image stylization. 
\begin{table}[!t]
\centering
\caption{\normalsize \textbf{User preference scores.}}
\vspace{-2mm}
\label{tbl:userstudy}
\resizebox{\columnwidth}{!}{
    \begin{tabular}{c|cccccc}
    \hline
    Metric & FreeStyle & StyleID & InstantStyle+ & Style-LoRA & DDS & Ours\\
    \hline
    Style & 0.060 & \underline{0.147} & 0.083 & 0.100 & 0.033 & \textbf{0.577} \\
    Content & 0.003 & 0.127 & \underline{0.136} & 0.090 & 0.017 & \textbf{0.627} \\
    Balance & 0.013 & 0.110 & \underline{0.127} & 0.077 & 0.020 & \textbf{0.653} \\
    \hline
    \end{tabular}
}
\vspace{-3mm}
\end{table}
\begin{figure}[!t]
    \centering
    \includegraphics[width=\linewidth]
    {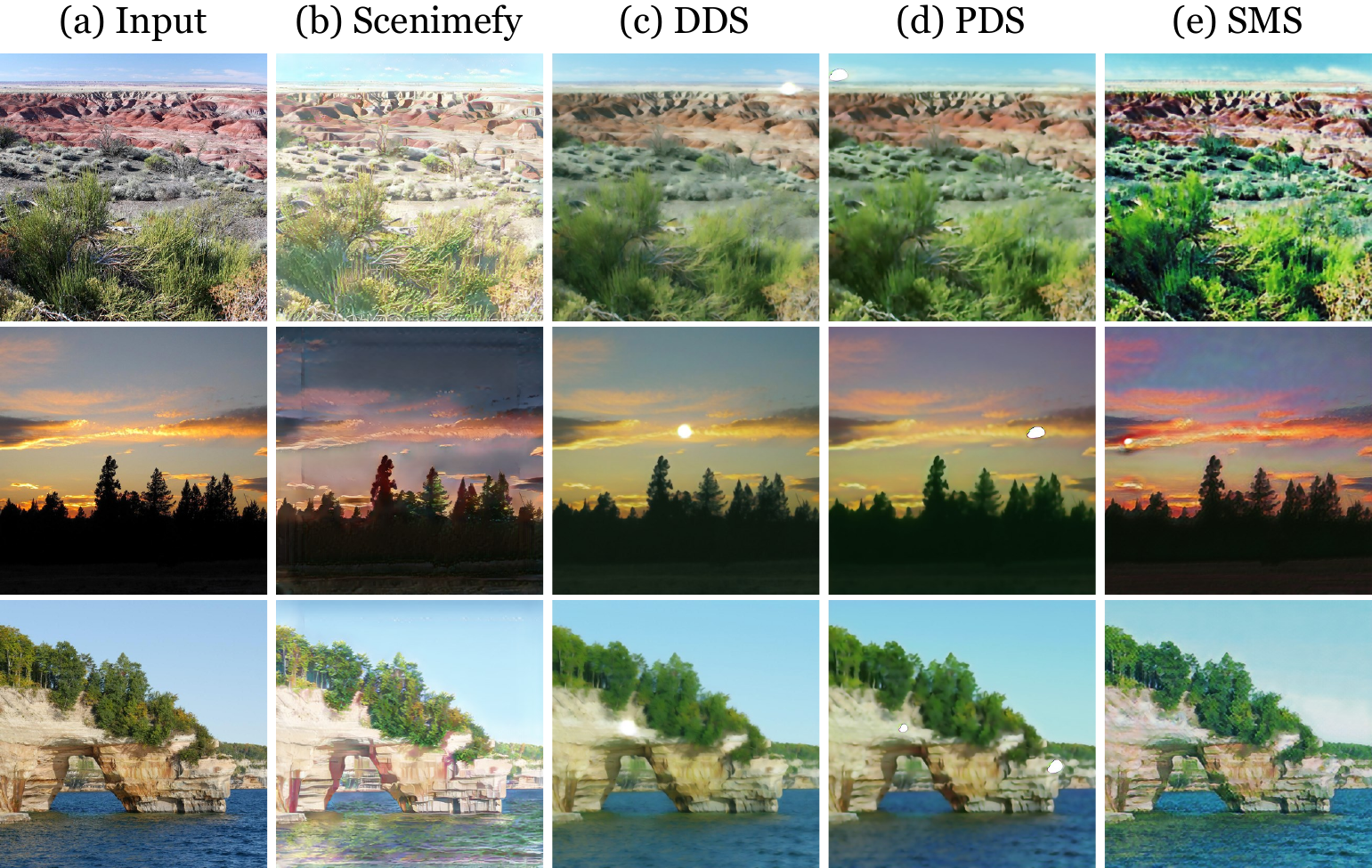}\vspace{-2mm}
    \caption{{\bf Feed-Forward stylization comparison.} }
    \label{fig:ffdresults} 
    \vspace{-2mm}
\end{figure}
\begin{table}[!t]
\centering
\caption{\textbf{Feed-Forward quantitative comparison.} The best results are in \textbf{bold}, and the second-best are \underline{underlined}.
}
\vspace{-2mm}
\label{tbl:ffd}
\resizebox{\columnwidth}{!}{
    \begin{tabular}{c|c|ccccc}
    \hline
    Metric & Real & Scenimefy~\cite{jiang2023scenimefy} & DDS~\cite{hertz2023delta} &  PDS~\cite{koo2024posterior} & SMS \\
    \hline
    LPIPS $\downarrow$ & 0.000 & 0.422 & \underline{0.321} & 0.427 & \textbf{0.268} \\
    FID $\downarrow$ & 12.608 & \textbf{12.054} & 12.886 & 16.229 & \underline{12.472} \\
    ArtFID $\downarrow$ & 13.608 & 18.561 & \underline{18.338} & 24.590 & \textbf{17.079} \\
    \hline
    \end{tabular}
}
\vspace{-5mm}
\end{table}
\begin{figure*}[!t]
    \centering
    \vspace{-5.5mm}
    \includegraphics[width=\linewidth]
    {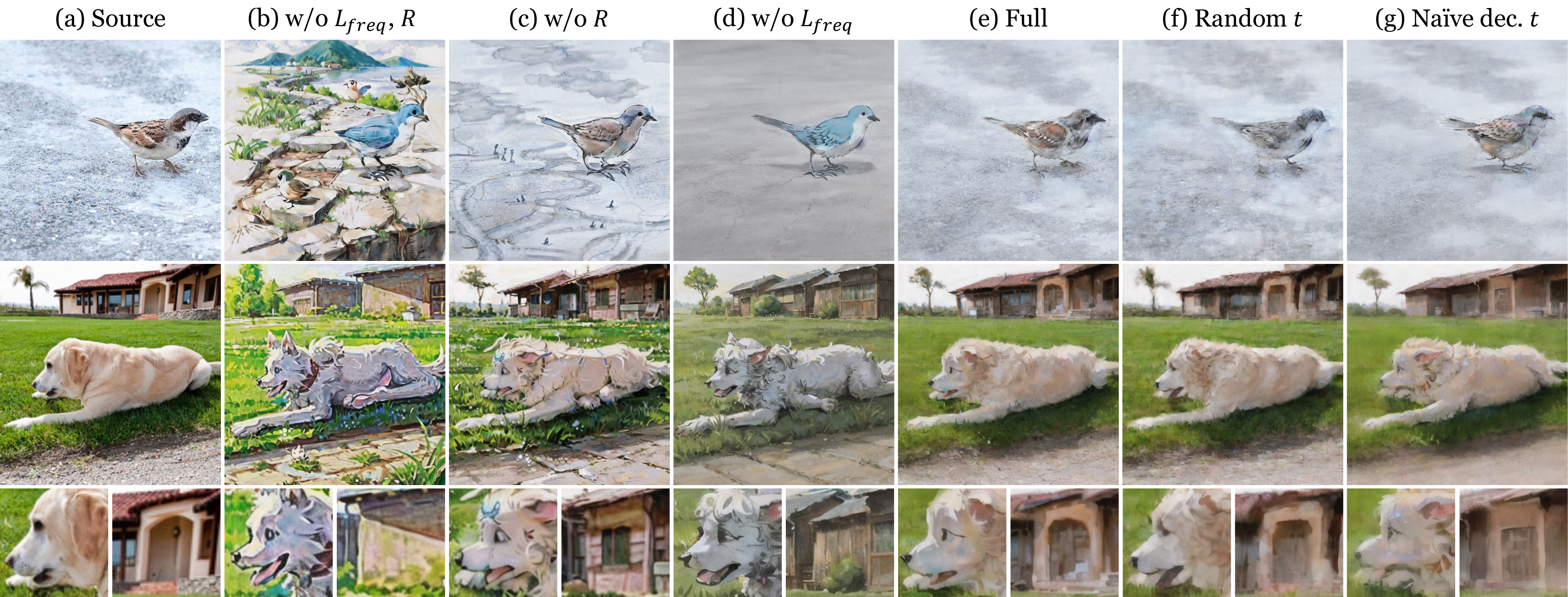}
    \vspace{-5mm}
    \caption{{\bf Ablation studies of SMS.} The effect of each key component is illustrated.}
    \vspace{-4.5mm}
    \label{fig:ablation}    
\end{figure*}

\noindent \textbf{Quantitative Results.}
Table~\ref{tbl:combined} presents a quantitative evaluation of our method against baselines.
For reference, we include metrics computed between the content and respective style datasets, labeled as ``Real''.
Our method surpasses baselines in context preservation, with significantly lower LPIPS and CFDS. 
Regarding style representation, SMS attains competitive FID scores and secures second-best results in PickScore and HPS, indicating strong visual appeal. 
While Style-LoRA performs slightly better in certain style metrics, this is expected, as our target style distribution aligns with Style-LoRA's generation distribution, serving as the upper bound for our method. 
Despite this, SMS effectively balances content and style, achieving the best ArtFID score across intricate and coarse artistic styles, further showcasing its robustness and versatility.
%

\noindent{\bf User Preference Score.} A total of $300$ comparisons from $20$ participants over $15$ cases across $6$ styles are collected.
Table~\ref{tbl:userstudy} summarizes the average preference scores, where we achieve the highest ratings in all three criteria, fruther suggesting the effectiveness of our method. Moreover, the results confirm that users prefer outcomes with better alignment to original content. 


\subsection{Feed-Forward Stylization}
\label{sec:ffd-exp}
\noindent \textbf{Experimental Settings.}
We train a lightweight feed-forward network $G_\theta$ ($\sim$43MB) using SMS for real-time inference and compare it with state-of-the-art GAN-based anime stylization methods like Scenimefy~\cite{jiang2023scenimefy}, and diffusion-based score optimization methods such as DDS~\cite{hertz2023delta} and PDS~\cite{koo2024posterior}. 
%
Our SMS setup follows the single-image experiments in Section~\ref{sec:single}, using a batch size $B$ of $4$.
%
We train on 60k LHQ~\cite{skorokhodov2021aligning} natural scenes at $512 \times512$ resolution, with a test set of $1,000$ sampled images from the remaining dataset.

\noindent \textbf{Qualitative Comparison.}
Figure~\ref{fig:ffdresults} shows that while Scenimefy produces visually pleasing results by leveraging real style data, it suffers from artifacts due to the instability inherent in GAN training. 
DDS and PDS, reliant on text-driven prompts, fail to achieve effective style transfer, leading to blurry outputs and weak style alignment. 
In contrast, SMS successfully distills styles into the model, capturing detailed stylistic features with fewer artifacts.
This stability is likely attributed to our score-based loss, which provides a more stable training process than GAN-based methods.

\noindent \textbf{Quantitative Results.}
Table~\ref{tbl:ffd} shows that our method achieves the lowest LPIPS and ArtFID scores, indicating superior content preservation and overall image quality. 
It also attains the second-best FID score, reflecting strong style adherence, second only to Scenimefy, which benefits from direct access to real anime data. 
%
%

%
\subsection{Ablation Studies}
\label{sec:abl-exp}
\noindent \textbf{SMS Components.} 
We systematically ablate each component in SMS to assess their contributions, with results shown in Figure~\ref{fig:ablation}.
%
Without any identity regularization (\textit{i.e.,} only the style matching objective), it successfully captures the overall Ghibli style but introduces noise and spurious details, such as background clutter in  Figure~\ref{fig:ablation}(Row~1) (b)).
Applying spectrum regularization $L_{freq}$ improves structure integrity and color alignment with the source image but leaves some high-frequency artifacts (see Figure~\ref{fig:ablation}(c)).
Semantic-aware gradient refinement $R$ selectively stylizes important regions, reducing disharmonious details and better balancing style and content (see Figure~\ref{fig:ablation}(d)). However, due to the absence of direct regularization with the source image, content distortion and color mismatches still exist.
Our full formulation, combining both components, achieves superior stylization with detailed textures, consistent colors, and strong content preservation (see Figure~\ref{fig:ablation}(e)).  
All the terms work together to improve the overall performance of the proposed style matching loss.
%


\noindent \textbf{Timestep Sampling Strategy.}
We compare our adaptive narrowing sampling strategy with random and naive linear decreasing sampling under identical settings.
Random sampling causes blurriness (see Figure~\ref{fig:ablation}(f)), while naive linear decreasing sampling fails to preserve local identity, such as a red scard-like artifact appearing around the dog's neck (see Figure~\ref{fig:ablation}(g)). In contrast, our method produces sharp, detailed stylization while preserving content fidelity.
%
%
The quantitative results in Table~\ref{tbl:quantabl} clearly demonstrate the effectiveness of each proposed module. 
Additional ablation studies are provided in the appendix.


\begin{table}[!t]
\centering
\caption{\normalsize \textbf{Quantitative ablation studies.}}
\vspace{-2mm}
\label{tbl:quantabl}
\resizebox{\columnwidth}{!}{
    \begin{tabular}{c|cccccc}
    \hline
    Metric & w/o $L_{f}$, R & w/o R & w/o $L_{f}$ &   Full & Random t & Na\"ive t\\
    \hline
    LPIPS $\downarrow$ & 0.703 & 0.505 & 0.536 & \textbf{0.326} & \underline{0.389} & 0.408 \\
    CFSD $\downarrow$ & 0.390 & 0.166 & 0.149 & \textbf{0.102} & \underline{0.107} & 0.118 \\
    FID $\downarrow$ & \underline{14.500} & 15.030 & 16.914 & \textbf{13.089} & 22.766 & 15.705 \\
    ArtFID $\downarrow$ & 26.403 & 24.132 & 27.514 & \textbf{18.686} & 32.936 & \underline{23.524} \\
    \hline
    \end{tabular}
}
\vspace{-5mm}
\end{table}
\section{Conclusion}
\label{sec:conlusion}

We propose Style Matching Score (SMS), an optimization method that formulates image stylization as a style distribution matching problem using style-specific LoRAs integrated with diffusion models.
%
Our approach balances style transfer and content preservation through progressive spectrum regularization and semantic-aware gradient refinement. 
%
SMS also demonstrates its versatility by distilling style into lightweight feed-forward networks.
%
Empirical results show its effectiveness in preserving content while adhering to target styles. 
%
Exploring SMS in broader parametric spaces, such as Neural Radiance Fields and 3D Gaussian Splatting can be interesting future work.

\noindent{\bf Acknowledgment.} This study is supported by the Ministry of Education, Singapore, under the Academic Research Fund Tier 1 (FY2023).
Yuxin Jiang is supported by the A*STAR ACIS Scholarship.

{
    \small
    \bibliographystyle{ieeenat_fullname}
    \bibliography{main}

\begin{thebibliography}{42}
\providecommand{\natexlab}[1]{#1}
\providecommand{\url}[1]{\texttt{#1}}
\expandafter\ifx\csname urlstyle\endcsname\relax
  \providecommand{\doi}[1]{doi: #1}\else
  \providecommand{\doi}{doi: \begingroup \urlstyle{rm}\Url}\fi

\bibitem[civ()]{civitai}
{Civit AI, Inc.}
\newblock \url{https://civitai.com/}.

\bibitem[Agustsson and Timofte(2017)]{agustsson2017ntire}
Eirikur Agustsson and Radu Timofte.
\newblock {NTIRE} 2017 challenge on single image super-resolution: Dataset and study.
\newblock In \emph{CVPR workshops}, 2017.

\bibitem[Chen et~al.(2018)Chen, Lai, and Liu]{chen2018cartoongan}
Yang Chen, Yu-Kun Lai, and Yong-Jin Liu.
\newblock Cartoon{GAN}: Generative adversarial networks for photo cartoonization.
\newblock In \emph{CVPR}, 2018.

\bibitem[Chung et~al.(2024)Chung, Hyun, and Heo]{Chung_2024_CVPR}
Jiwoo Chung, Sangeek Hyun, and Jae-Pil Heo.
\newblock Style injection in diffusion: A training-free approach for adapting large-scale diffusion models for style transfer.
\newblock In \emph{CVPR}, 2024.

\bibitem[Everaert et~al.(2023)Everaert, Bocchio, Arpa, S{\"u}sstrunk, and Achanta]{everaert2023diffusion}
Martin~Nicolas Everaert, Marco Bocchio, Sami Arpa, Sabine S{\"u}sstrunk, and Radhakrishna Achanta.
\newblock Diffusion in style.
\newblock In \emph{ICCV}, 2023.

\bibitem[Gatys et~al.(2016)Gatys, Ecker, and Bethge]{gatys2016image}
Leon~A Gatys, Alexander~S Ecker, and Matthias Bethge.
\newblock Image style transfer using convolutional neural networks.
\newblock In \emph{CVPR}, 2016.

\bibitem[He et~al.(2024)He, Li, Zhang, Yan, Si, Li, and Shen]{he2024freestyle}
Feihong He, Gang Li, Mengyuan Zhang, Leilei Yan, Lingyu Si, Fanzhang Li, and Li Shen.
\newblock {FreeStyle}: Free lunch for text-guided style transfer using diffusion models.
\newblock \emph{arXiv preprint arXiv:2401.15636}, 2024.

\bibitem[Hertz et~al.(2023)Hertz, Aberman, and Cohen-Or]{hertz2023delta}
Amir Hertz, Kfir Aberman, and Daniel Cohen-Or.
\newblock {Delta Denoising Score}.
\newblock In \emph{ICCV}, 2023.

\bibitem[Heusel et~al.(2017)Heusel, Ramsauer, Unterthiner, Nessler, and Hochreiter]{heusel2017gans}
Martin Heusel, Hubert Ramsauer, Thomas Unterthiner, Bernhard Nessler, and Sepp Hochreiter.
\newblock {GANs} trained by a two time-scale update rule converge to a local nash equilibrium.
\newblock \emph{NeurIPS}, 2017.

\bibitem[Ho et~al.(2020)Ho, Jain, and Abbeel]{ho2020denoising}
Jonathan Ho, Ajay Jain, and Pieter Abbeel.
\newblock Denoising diffusion probabilistic models.
\newblock \emph{NeurIPS}, 2020.

\bibitem[H{\"o}llein et~al.(2022)H{\"o}llein, Johnson, and Nie{\ss}ner]{hollein2022stylemesh}
Lukas H{\"o}llein, Justin Johnson, and Matthias Nie{\ss}ner.
\newblock {StyleMesh}: Style transfer for indoor 3d scene reconstructions.
\newblock In \emph{CVPR}, 2022.

\bibitem[Hu et~al.(2022)Hu, Shen, Wallis, Allen-Zhu, Li, Wang, Wang, and Chen]{hu2022lora}
Edward~J Hu, Yelong Shen, Phillip Wallis, Zeyuan Allen-Zhu, Yuanzhi Li, Shean Wang, Lu Wang, and Weizhu Chen.
\newblock Lo{RA}: Low-rank adaptation of large language models.
\newblock In \emph{ICLR}, 2022.

\bibitem[Huang and Belongie(2017)]{huang2017arbitrary}
Xun Huang and Serge Belongie.
\newblock Arbitrary style transfer in real-time with adaptive instance normalization.
\newblock In \emph{ICCV}, 2017.

\bibitem[Jiang et~al.(2023)Jiang, Jiang, Yang, and Loy]{jiang2023scenimefy}
Yuxin Jiang, Liming Jiang, Shuai Yang, and Chen~Change Loy.
\newblock Scenimefy: learning to craft anime scene via semi-supervised image-to-image translation.
\newblock In \emph{ICCV}, 2023.

\bibitem[Johnson et~al.(2016)Johnson, Alahi, and Fei-Fei]{Johnson2016Perceptual}
Justin Johnson, Alexandre Alahi, and Li Fei-Fei.
\newblock Perceptual losses for real-time style transfer and super-resolution.
\newblock In \emph{ECCV}, 2016.

\bibitem[Ju et~al.(2024)Ju, Zeng, Bian, Liu, and Xu]{ju2023direct}
Xuan Ju, Ailing Zeng, Yuxuan Bian, Shaoteng Liu, and Qiang Xu.
\newblock {PnP Inversion}: Boosting diffusion-based editing with 3 lines of code.
\newblock \emph{ICLR}, 2024.

\bibitem[Kang et~al.(2024)Kang, Zhang, Barnes, Paris, Kwak, Park, Shechtman, Zhu, and Park]{kang2024diffusion2gan}
Minguk Kang, Richard Zhang, Connelly Barnes, Sylvain Paris, Suha Kwak, Jaesik Park, Eli Shechtman, Jun-Yan Zhu, and Taesung Park.
\newblock {Distilling Diffusion Models into Conditional GANs}.
\newblock In \emph{ECCV}, 2024.

\bibitem[Karras et~al.(2022)Karras, Aittala, Aila, and Laine]{karras2022elucidating}
Tero Karras, Miika Aittala, Timo Aila, and Samuli Laine.
\newblock Elucidating the design space of diffusion-based generative models.
\newblock \emph{NeurIPS}, 2022.

\bibitem[Kirstain et~al.(2023)Kirstain, Polyak, Singer, Matiana, Penna, and Levy]{kirstain2023pick}
Yuval Kirstain, Adam Polyak, Uriel Singer, Shahbuland Matiana, Joe Penna, and Omer Levy.
\newblock Pick-a-{P}ic: An open dataset of user preferences for text-to-image generation.
\newblock \emph{NeurIPS}, 2023.

\bibitem[Koo et~al.(2024)Koo, Park, and Sung]{koo2024posterior}
Juil Koo, Chanho Park, and Minhyuk Sung.
\newblock {Posterior Distillation Sampling}.
\newblock In \emph{CVPR}, 2024.

\bibitem[McAllister et~al.(2024)McAllister, Ge, Huang, Jacobs, Efros, Holynski, and Kanazawa]{mcallister2024rethinking}
David McAllister, Songwei Ge, Jia-Bin Huang, David~W. Jacobs, Alexei~A. Efros, Aleksander Holynski, and Angjoo Kanazawa.
\newblock Rethinking score distillation as a bridge between image distributions.
\newblock \emph{NeurIPS}, 2024.

\bibitem[Mirzaei et~al.(2024)Mirzaei, Aumentado-Armstrong, Brubaker, Kelly, Levinshtein, Derpanis, and Gilitschenski]{mirzaei2023watchyoursteps}
Ashkan Mirzaei, Tristan Aumentado-Armstrong, Marcus~A. Brubaker, Jonathan Kelly, Alex Levinshtein, Konstantinos~G. Derpanis, and Igor Gilitschenski.
\newblock Watch your steps: Local image and scene editing by text instructions.
\newblock In \emph{ECCV}, 2024.

\bibitem[Nam et~al.(2024)Nam, Kwon, Park, and Ye]{nam2024contrastive}
Hyelin Nam, Gihyun Kwon, Geon~Yeong Park, and Jong~Chul Ye.
\newblock Contrastive denoising score for text-guided latent diffusion image editing.
\newblock In \emph{CVPR}, 2024.

\bibitem[Poole et~al.(2023)Poole, Jain, Barron, and Mildenhall]{pooledreamfusion}
Ben Poole, Ajay Jain, Jonathan~T Barron, and Ben Mildenhall.
\newblock {DreamFusion}: Text-to-3d using 2d diffusion.
\newblock In \emph{ICLR}, 2023.

\bibitem[Radford et~al.(2021)Radford, Kim, Hallacy, Ramesh, Goh, Agarwal, Sastry, Askell, Mishkin, Clark, et~al.]{radford2021learning}
Alec Radford, Jong~Wook Kim, Chris Hallacy, Aditya Ramesh, Gabriel Goh, Sandhini Agarwal, Girish Sastry, Amanda Askell, Pamela Mishkin, Jack Clark, et~al.
\newblock Learning transferable visual models from natural language supervision.
\newblock In \emph{ICML}, 2021.

\bibitem[Rombach et~al.(2022)Rombach, Blattmann, Lorenz, Esser, and Ommer]{rombach2022high}
Robin Rombach, Andreas Blattmann, Dominik Lorenz, Patrick Esser, and Bj{\"o}rn Ommer.
\newblock High-resolution image synthesis with latent diffusion models.
\newblock In \emph{CVPR}, 2022.

\bibitem[Ruiz et~al.(2024)Ruiz, Li, Wadhwa, Pritch, Rubinstein, Jacobs, and Fruchter]{ruiz2024magic}
Nataniel Ruiz, Yuanzhen Li, Neal Wadhwa, Yael Pritch, Michael Rubinstein, David~E Jacobs, and Shlomi Fruchter.
\newblock Magic insert: Style-aware drag-and-drop.
\newblock \emph{arXiv preprint arXiv:2407.02489}, 2024.

\bibitem[Saharia et~al.(2022)Saharia, Chan, Chang, Lee, Ho, Salimans, Fleet, and Norouzi]{saharia2022palette}
Chitwan Saharia, William Chan, Huiwen Chang, Chris Lee, Jonathan Ho, Tim Salimans, David Fleet, and Mohammad Norouzi.
\newblock Palette: Image-to-image diffusion models.
\newblock In \emph{ACM SIGGRAPH 2022 conference proceedings}, 2022.

\bibitem[Skorokhodov et~al.(2021)Skorokhodov, Sotnikov, and Elhoseiny]{skorokhodov2021aligning}
Ivan Skorokhodov, Grigorii Sotnikov, and Mohamed Elhoseiny.
\newblock Aligning latent and image spaces to connect the unconnectable.
\newblock In \emph{ICCV}, 2021.

\bibitem[Song et~al.(2021{\natexlab{a}})Song, Meng, and Ermon]{songdenoising}
Jiaming Song, Chenlin Meng, and Stefano Ermon.
\newblock Denoising diffusion implicit models.
\newblock In \emph{ICLR}, 2021{\natexlab{a}}.

\bibitem[Song et~al.(2021{\natexlab{b}})Song, Sohl-Dickstein, Kingma, Kumar, Ermon, and Poole]{songscore}
Yang Song, Jascha Sohl-Dickstein, Diederik~P Kingma, Abhishek Kumar, Stefano Ermon, and Ben Poole.
\newblock Score-based generative modeling through stochastic differential equations.
\newblock In \emph{ICLR}, 2021{\natexlab{b}}.

\bibitem[Song et~al.(2025)Song, Liu, and Shou]{song2025omniconsistency}
Yiren Song, Cheng Liu, and Mike~Zheng Shou.
\newblock Omniconsistency: Learning style-agnostic consistency from paired stylization data.
\newblock \emph{arXiv preprint arXiv:2505.18445}, 2025.

\bibitem[Wang et~al.(2024)Wang, Xing, Huang, Ai, Wang, and Bai]{wang2024instantstyle}
Haofan Wang, Peng Xing, Renyuan Huang, Hao Ai, Qixun Wang, and Xu Bai.
\newblock {InstantStyle-Plus}: Style transfer with content-preserving in text-to-image generation.
\newblock \emph{arXiv preprint arXiv:2407.00788}, 2024.

\bibitem[Wang et~al.(2023)Wang, Lu, Wang, Bao, Li, Su, and Zhu]{wang2023prolificdreamer}
Zhengyi Wang, Cheng Lu, Yikai Wang, Fan Bao, Chongxuan Li, Hang Su, and Jun Zhu.
\newblock {ProlificDreamer}: High-fidelity and diverse text-to-3d generation with variational score distillation.
\newblock In \emph{NeurIPS}, 2023.

\bibitem[Wright and Ommer(2022)]{wright2022artfid}
Matthias Wright and Bj{\"o}rn Ommer.
\newblock {ArtFID}: Quantitative evaluation of neural style transfer.
\newblock In \emph{GCPR}, 2022.

\bibitem[Wu et~al.(2023)Wu, Hao, Sun, Chen, Zhu, Zhao, and Li]{wu2023human}
Xiaoshi Wu, Yiming Hao, Keqiang Sun, Yixiong Chen, Feng Zhu, Rui Zhao, and Hongsheng Li.
\newblock Human preference score v2: A solid benchmark for evaluating human preferences of text-to-image synthesis.
\newblock \emph{arXiv preprint arXiv:2306.09341}, 2023.

\bibitem[Yang et~al.(2023)Yang, Hwang, and Ye]{yang2023zero}
Serin Yang, Hyunmin Hwang, and Jong~Chul Ye.
\newblock Zero-shot contrastive loss for text-guided diffusion image style transfer.
\newblock In \emph{ICCV}, 2023.

\bibitem[Yin et~al.(2024)Yin, Gharbi, Zhang, Shechtman, Durand, Freeman, and Park]{yin2024onestep}
Tianwei Yin, Micha{\"e}l Gharbi, Richard Zhang, Eli Shechtman, Fr{\'e}do Durand, William~T Freeman, and Taesung Park.
\newblock One-step diffusion with distribution matching distillation.
\newblock In \emph{CVPR}, 2024.

\bibitem[Zhang et~al.(2023{\natexlab{a}})Zhang, Rao, and Agrawala]{zhang2023adding}
Lvmin Zhang, Anyi Rao, and Maneesh Agrawala.
\newblock Adding conditional control to text-to-image diffusion models.
\newblock In \emph{ICCV}, 2023{\natexlab{a}}.

\bibitem[Zhang et~al.(2018)Zhang, Isola, Efros, Shechtman, and Wang]{zhang2018unreasonable}
Richard Zhang, Phillip Isola, Alexei~A Efros, Eli Shechtman, and Oliver Wang.
\newblock The unreasonable effectiveness of deep features as a perceptual metric.
\newblock In \emph{CVPR}, 2018.

\bibitem[Zhang et~al.(2023{\natexlab{b}})Zhang, Huang, Tang, Huang, Ma, Dong, and Xu]{zhang2023inversion}
Yuxin Zhang, Nisha Huang, Fan Tang, Haibin Huang, Chongyang Ma, Weiming Dong, and Changsheng Xu.
\newblock Inversion-based style transfer with diffusion models.
\newblock In \emph{CVPR}, 2023{\natexlab{b}}.

\bibitem[Zhu et~al.(2017)Zhu, Park, Isola, and Efros]{zhu2017unpaired}
Jun-Yan Zhu, Taesung Park, Phillip Isola, and Alexei~A Efros.
\newblock Unpaired image-to-image translation using cycle-consistent adversarial networks.
\newblock In \emph{ICCV}, 2017.

\end{thebibliography}
}

\clearpage
\section*{Appendix}
\label{sec:appendix}

The document provides supplementary information not elaborated on in our main paper due to space constraints: implementation details (Section~\ref{sec:implement}), derivation of the Style Matching Objective (Section~\ref{sec:derivation}), spectrum-based style analysis (Section~\ref{sec:rapsd}), additional optimization-based comparisons (Section~\ref{sec:optimization}), further studies (Section~\ref{sec:studies}), more qualitative comparisons (Section~\ref{sec:comparison}), additional results (Section~\ref{sec:results}), limitations (Section~\ref{sec:limitations}), and broader impact (Section~\ref{sec:impact}). {Code: \url{https://github.com/showlab/SMS}}.

\appendix
\begin{figure*}[!ht]
    \centering
    \includegraphics[width=\linewidth]{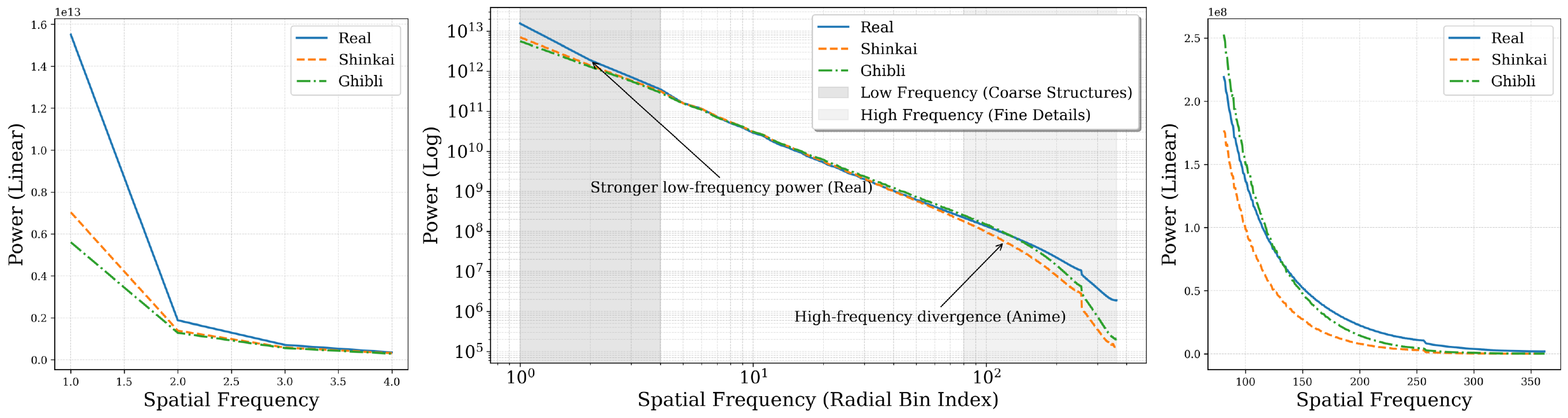}
    \vspace{-5.5mm}
    \caption{{\bf Comparison of RAPSD}: Real-world images (Real) vs. Anime styles (Shinkai and Ghibli). \textbf{Left:} Zoom-in on low-frequency range (linear scale). \textbf{Middle:} Full-spectrum analysis (log-log scale). \textbf{Right:} Zoom-in on the high-frequency range (linear scale), revealing reduced high-frequency power in anime styles, which corresponds to smoother textures and stylized simplicity.}
    \vspace{-3.5mm}
    \label{fig:rapsd}
\end{figure*} 

\section{Implementation Details}
\label{sec:implement}
\subsection{SMS Training Procedure}
Algorithm~\ref{alg:distillation} details our style matching training procedure. 
%

\subsection{Style Data}
We leverage the off-the-shelf style LoRA from Civitai~\cite{civitai} to support diverse artistic styles.
To ensure fairness when comparing with baseline methods that utilize different style representations, we carefully adopt the following adaptations: 1) Text-driven (\textit{e.g.}, FreeStyle~\cite{he2024freestyle}, DDS~\cite{hertz2023delta}): Use descriptive text prompts to capture the style. 
2) Exemplar-guided (\textit{e.g.}, StyleID~\cite{Chung_2024_CVPR}, InstantStyle-Plus~\cite{wang2024instantstyle}): Source reference images from training data. 
3) Collection-based (\textit{e.g.}, Style-LoRA): Use Exactly the same LoRA. 
\subsection{Training time}
Table~\ref{tbl:traintime} reports the per-image runtime (seconds) for baselines on an NVIDIA L40 GPU under default settings. 
Although some baselines require only forward steps, they require additional processing steps such as DDIM~\cite{songdenoising} inversion and substantial preparation (\textit{e.g.}, ControlNet~\cite{zhang2023adding} training for InstantStyle-Plus~\cite{wang2024instantstyle} and Style-LoRA). 
In contrast, our SMS runs without any model-specific preparation, achieving a comparable overall runtimes. 
Furthermore, the optimization-based nature of SMS enables it to extend to more complex, parameterized representations, which is not straightforward with other methods.
%

\begin{algorithm}[!h]
    \footnotesize
    \caption{\label{alg:distillation}SMS Training Procedure}
    \KwIn{Source image $x^\text{src}$; text prompt $y^\text{src}$; editing instruction $y^\text{edit}$; number of training iterations $N$}
    \KwOut{Trained generator $G_\theta$}
    \KwRequire{Pretrained SD diffusion denoiser $\epsilon_\text{real}$; style-specific LoRA integrated into $\epsilon_\text{real}$ yielding $\epsilon_\text{style}$; trainable LoRA integrated into $\epsilon_\text{real}$ yielding $\epsilon_\text{fake}^\phi$; SD VAE encoder $\mathcal{E}$}
    \KwInitialize{$\epsilon_\text{fake}^\phi \leftarrow \text{copyWeights}(\epsilon_\text{real})$}
        \For{$\text{i} = 1$ \KwTo $N$}{
        \tcc{Generate stylized images}
        $x^\text{tgt} \leftarrow G_\theta(x^\text{src})$
        
        \text{~}

        \tcc{Prepare latents}
        $z_0^\text{src} \leftarrow \mathcal{E}(x^\text{src})$
        
        $z_0^\text{tgt} \leftarrow \mathcal{E}(x^\text{tgt})$
        
        \tcp{Adaptive Narrowing Sampling}
        
        Sample $t \sim \mathcal{U}(t_\text{min}, t_\text{upper})$ 
        
        Sample $\epsilon \sim \mathcal{N}(0, \mathbf{I})$

        $z_{t}^\text{src} \leftarrow \sqrt{\bar{\alpha}_t} z_{0}^\text{src} + \sqrt{1 - \bar{\alpha}_t} \epsilon$
        
        $z_{t}^\text{tgt} \leftarrow \sqrt{\bar{\alpha}_t} z_{0}^\text{tgt} + \sqrt{1 - \bar{\alpha}_t} \epsilon$

        \text{~}
        
        \tcc{Update generator}
        
        \tcp{Semantic-Aware Gradient Refinement}
        
        $\mathcal{R}(z_t^{\text{src}}, t) = \text{Norm} ( | \epsilon_{\text{real}}(z_t^{\text{src}}; y_{\text{edit}}, t) - \epsilon_{\text{real}}(z_t^{\text{src}}; y_{\emptyset}, t) | )$ 

        $\mathcal{L}_\text{style} \leftarrow ||\mathcal{R} \odot [w_t( \epsilon_{\text{style}}(z_t^\text{tgt}; y_\text{src}, t)  - \epsilon_\text{fake}^\phi(z_t^\text{tgt}; y_\text{src}, t))]||_2^2$  

        \tcp{Progressive Spectrum Reglarization}
        
        $\mathcal{L}_\text{freq} \leftarrow ||\mathcal{F}_\text{low}(z_0^\text{tgt}, t), \mathcal{F}_\text{low}(z_0^\text{src}, t)||_2^2$
        
        $\mathcal{L}_\text{SMS} \leftarrow \mathcal{L}_\text{style} + \lambda\mathcal{L}_\text{freq}$
        
        $G_\theta \leftarrow \text{update}(\theta, \nabla_\theta \mathcal{L}_\text{SMS})$
        
        \text{~}
        
        \tcc{Update trainable LoRA}
        Sample $t\sim\mathcal{U}(t_\text{min}, t_\text{max})$

        Sample $\epsilon \sim \mathcal{N}(0, \mathbf{I})$
        
        $z_{t}^\text{tgt} \leftarrow \sqrt{\bar{\alpha}_t} z_{0}^\text{tgt} + \sqrt{1 - \bar{\alpha}_t} \epsilon$
        
        $ \mathcal{L}_{\text{denoise}}^\phi \leftarrow ||\epsilon_\text{fake}^\phi(z_t^\text{tgt}, t) - \epsilon||_2^2$
        
        $\epsilon_\text{fake}^\phi \leftarrow \text{update}(\phi, \nabla_\phi \mathcal{L}_{\text{denoise}}^\phi)$
    }
\end{algorithm}

\begin{table}[!h]
\centering
\vspace{-2mm}
\caption{Image stylization per-image runtime comparison.}
\label{tbl:traintime}
\vspace{-1mm}
\resizebox{\columnwidth}{!}{
    \begin{tabular}{c|cccccc}
    \hline
     & FreeStyle & StyleID & InstantStyle+ & Style-LoRA & DDS & SMS\\
    \hline
    Style & Text & Exemplar & Exemplar & LoRA & Text & LoRA \\
    \multirow{2}{*}{Train} & - & - & ControlNet ($\sim600$ h) + & ControlNet & - & -  \\
    &   &   & IPAdapter ($\sim192$ h) & ($\sim600$ h)  &   &   \\
    DDIM Inv & - & 6.553 & 23.688 & - & - & -  \\
    Inference & 28.136 & 2.683 & 18.375 & 2.323 & 31.716 & 87.582 \\
    \hline
    \end{tabular}
}
\vspace{-4mm}
\end{table}

\section{Derivation for Style Matching Objective}
\label{sec:derivation}
We derive the style matching objective (see Section~$3.2$ in the main paper) by using score functions approximated by DMs to minimize the KL divergence between the generated distribution $p_{G_\theta}$ and the target style distribution $p_\text{style}$. This derivation connects Equation~$(1)$ to Equation~$(2)$ in the main paper.

\subsection{Gradient of the KL Divergence}

Starting from the KL divergence:
\begin{equation} 
\label{eq:kl1} 
D_{\text{KL}}(p_{G_\theta} || p_{\text{style}}) = \int p_{G_\theta}(x^{\text{tgt}}) \log \frac{p_{G_\theta}(x^{\text{tgt}})}{p_{\text{style}}(x^{\text{tgt}})} dx^{\text{tgt}}, 
\end{equation}
where $x^{\text{tgt}} = G_\theta(x^\text{src})$ and $G_\theta$ is the generator parametrized by $\theta$.
Our goal is to compute the gradient of $D_\text{KL}$ with respect to $\theta$:
\begin{equation} 
\label{eq:gradientkl} 
\nabla_\theta D_{\text{KL}} = \nabla_\theta \int p_{G_\theta}(x^{\text{tgt}}) \log \frac{p_{G_\theta}(x^{\text{tgt}})}{p_{\text{style}}(x^{\text{tgt}})} dx^{\text{tgt}}. \end{equation}
Using the property $\nabla_\theta p_{G_\theta}(x) = p_{G_\theta}(x) \nabla_x \log p_{G_\theta}(x)$, we can express the gradient as:
\begin{equation} 
\label{eq:gradientkl2} 
\nabla_\theta D_{\text{KL}} = \int p_{G_\theta}(x) \nabla_\theta \log p_{G_\theta}(x) \log \frac{p_{G_\theta}(x^{\text{tgt}})}{p_{\text{style}}(x^{\text{tgt}})}dx^{\text{tgt}}. \end{equation}
Since $p_{\text{style}}$ does not depend on $\theta$, we have $\nabla_\theta \log p_{\text{style}}(x^{\text{tgt}}) = 0$. 
Furthermore, using the chain rule, we can compute $\nabla_\theta \log p_{G_\theta}(x^{\text{tgt}})$ as follows:
\begin{equation} 
\label{eq:1} 
\begin{aligned}
\nabla_\theta \log p_{G_\theta}(x^{\text{tgt}}) 
&= \left( \nabla_{x^{\text{tgt}}} \log p_{G_\theta}(x^{\text{tgt}}) \right) \frac{\partial x^{\text{tgt}}}{\partial \theta} \\
&= s_{G_\theta}(x^{\text{tgt}}) \frac{\partial G_\theta(x^{\text{src}})}{\partial \theta},
\end{aligned}
\end{equation}
where $s_{G_\theta}(x) := s_\text{fake}(x) = \nabla_x \log p_{G_\theta}(x)$ is the score function of the generated distribution. Following DMD~\cite{yin2024onestep}, we name it the fake score.
Substituting back into Equation~(\ref{eq:gradientkl2}):
\begin{equation} 
\nabla_\theta D_{\text{KL}} = \int p_{G_\theta}(x^{\text{tgt}}) s_\text{fake}(x^{\text{tgt}}) \log \frac{p_{G_\theta}(x^{\text{tgt}})}{p_{\text{style}}(x^{\text{tgt}})} \frac{\partial G_\theta(x^{\text{src}})}{\partial \theta}. 
\end{equation}
Recognizing that the gradient of the log-density ratio is the difference of the score functions:
\begin{equation} 
\label{eq:2} 
\nabla_x \log \frac{p_{G_\theta}(x)}{p_{\text{style}}(x)} = s_\text{fake}(x) - s_{\text{style}}(x), 
\end{equation}
where $s_{\text{style}}(x) = \nabla_x \log p_\text{style}(x)$ is the score function of the target style distribution.
%
The integral can be expressed as an expectation over $x \sim p_{G_\theta}$:
\begin{equation} 
\label{eq:klscores} 
\nabla_\theta D_{\text{KL}} = \underset{x^{\text{tgt}} \sim p{G_\theta}}{\mathbb{E}} \left[\left(s_{\text{style}}(x^{\text{tgt}}) - s_\text{fake}(x^{\text{tgt}}) \right) \frac{\partial G_\theta(x^{\text{src}})}{\partial \theta} \right], 
\end{equation}
indicating that the gradient is pointing in the direction that moves $p_{G_\theta}$ closer to $p_\text{style}$.

\subsection{Approximating Score Functions with Diffusion Models}
We approximate the score functions $s_\text{style}(x^\text{tgt})$ and $s_\text{fake}(x^\text{tgt}$ using diffusion models~\cite{yin2024onestep, songscore}.
The score function of the data distribution $s(x)$ is related to the time-dependent score function $s(z_t, t)$ through the diffusion process, where $z_t$ is obtained by adding Gaussian noise to $z_0 = \mathcal{E}(x)$.

\noindent \textbf{Equivalence of Noise and Data Prediction}
Before proceeding with the substitution into the gradient expression, it is beneficial to convert the data prediction models $\mu$ to noise prediction models $\epsilon$.
This conversion simplifies the derivation and aligns with practical implementations, as DMs are typically trained to predict the noise.
The relationship is given by~\cite{karras2022elucidating}:
\begin{equation}
\label{eq:convert}
    \mu(z_t, t) =  \frac{z_t-\sigma_t\epsilon(z_t, t)}{\alpha_t}.
\end{equation}
Rewriting the score function in terms of the noise prediction model, we have:
\begin{equation} 
\label{eq:3} 
\begin{aligned}
s(z_t, t) = \nabla_{z_t} \log p(z_t) 
&= \frac{z_t - \alpha_t \mu(z_t, t)}{\sigma_t^2} =  \frac{\epsilon(z_t, t)}{\sigma_t}
\end{aligned}
\end{equation}

\noindent \textbf{Target style score.} The target style distribution $p_\text{style}(x)$ is modeled using a pretrained DM with a style-specific LoRA $\epsilon_\text{style}$. The score function is: $s_\text{style}(z_t, t) = \frac{\epsilon_{\text{style}}(z_t, t)}{\sigma_t}.$

\noindent \textbf{Generated fake score.}
Similarly, we model the generated distribution $p_{G_\theta}$ using a DM with trainable LoRA $\epsilon_{\text{fake}}^\phi$. The score function is: $s_\text{fake}(z_t, t) = \frac{\epsilon_{\text{fake}}^\phi(z_t, t)}{\sigma_t}.$
We train $\epsilon_{\text{fake}}^\phi$ to model the distribution of the generated images $z_0^\text{tgt} = \mathcal{E}(G_\theta(x^\text{src}))$ by minimizing the standard denoising objective~\cite{ho2020denoising}:
\begin{equation}
\label{eq:fake_train}
    \mathcal{L}_\text{denoise}^\phi = ||\epsilon_{\text{fake}}^\phi(z_t, t) - \epsilon||_2^2,
\end{equation}
Substituting the approximations into Equation~(\ref{eq:klscores}), we obtain:
\begin{equation} 
\label{eq:final1}
\nabla_\theta D_{\text{KL}} \simeq \underset{t,\epsilon}{\mathbb{E}} \left[ w_t \left(\epsilon_{\text{style}}(z_t, t) - \epsilon_{\text{fake}}^\phi(z_t, t)\right) \frac{\partial G_\theta(x^{\text{src}})}{\partial \theta} \right].
\end{equation}

\section{Spectrum-Based Style Analysis}
\label{sec:rapsd}
To identify and quantify the gap between the real and style domains, we analyze their spectral differences, focusing on two representative anime styles: Shinkai and Ghibli. 
Using $5,958$ Shinkai images~\cite{jiang2023scenimefy}, $714$ Ghibli images and $90,000$ real-world images~\cite{skorokhodov2021aligning}, we calculate the Radially Averaged Power Spectral Density (RAPSD) for each domain.

Figure~\ref{fig:rapsd} shows that real images have consistently higher power at both low and high frequencies. 
In contrast, anime styles demonstrate reduced high-frequency power, suggesting smoother textures and a uniform representation of details. This aligns with its artistic choices in anime, where sharp transitions and clean edges are emphasized while avoiding natural noise and irregularities in real-world images. 
Inspired by this gap, we propose a progressive spectrum regularization term (see Section $3.3$) that aligns the spectral properties of generated images with the target style domain, allowing faithful stylization while maintaining structural fidelity.

\section{Additional Optimization-based Method Comparisons}
\label{sec:optimization}
\begin{figure}[!t]
    \centering
    \vspace{-4mm}
    \includegraphics[width=\linewidth]
    {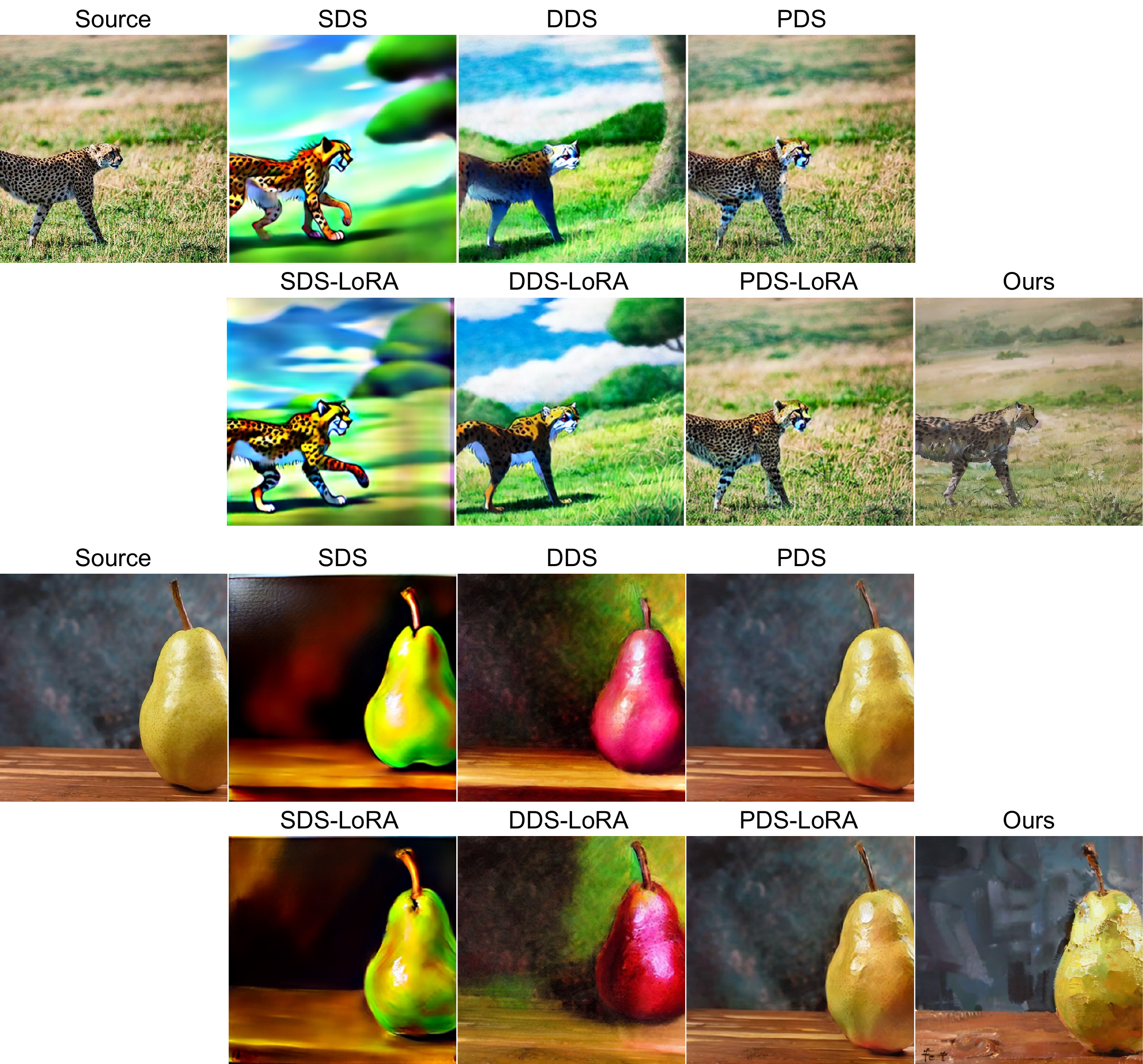}
    \caption{Comparison of optimization-based methods (SDS~\cite{pooledreamfusion}, DDS~\cite{hertz2023delta} and PDS~\cite{koo2024posterior}) with and without style LoRA priors on Ghibli and oil painting styles.}
    \label{fig:sds-comp} 
\end{figure}

In the main paper, we select DDS~\cite{hertz2023delta} as the representative optimization-based method for clarity.
Although other score distillation methods such as SDS~\cite{pooledreamfusion} and PDS~\cite{koo2024posterior} are technically relevant, our experiments show that these methods fail in global style transfer, resulting in poorer performance (see Figure~\ref{fig:sds-comp}(Row 1,3)).
Furthermore, when we apply the same style LoRA priors to these text-guided optimization methods, the results (see Figure~\ref{fig:sds-comp}(Row 2,4)) indicate that they do not fully leverage the style LoRA for capturing style information.
%

\section{Further Studies}
\label{sec:studies}
\subsection{Identity Loss Variant Study}
In Section~$3.3$ of the main paper, we introduce a novel progressive spectrum regularization in the frequency domain, instead of traditional spatial domain identity preservation losses. 
While we have already ablated its effectiveness in Section~$5.3$, we further verify its utility by comparing it against other latent space identity loss variants: spatial Mean Square Error (MSE)~\cite{koo2024posterior} and E-LatentLPIPS~\cite{kang2024diffusion2gan}.
Additionally, we test a fixed frequency threshold $\text{thld}(t) = 0.3$, retaining the top $30\%$ of low-frequency components, as opposed to our timestep-aware progressive approach.
\begin{figure}[t]
    \centering
    \includegraphics[width=\linewidth]{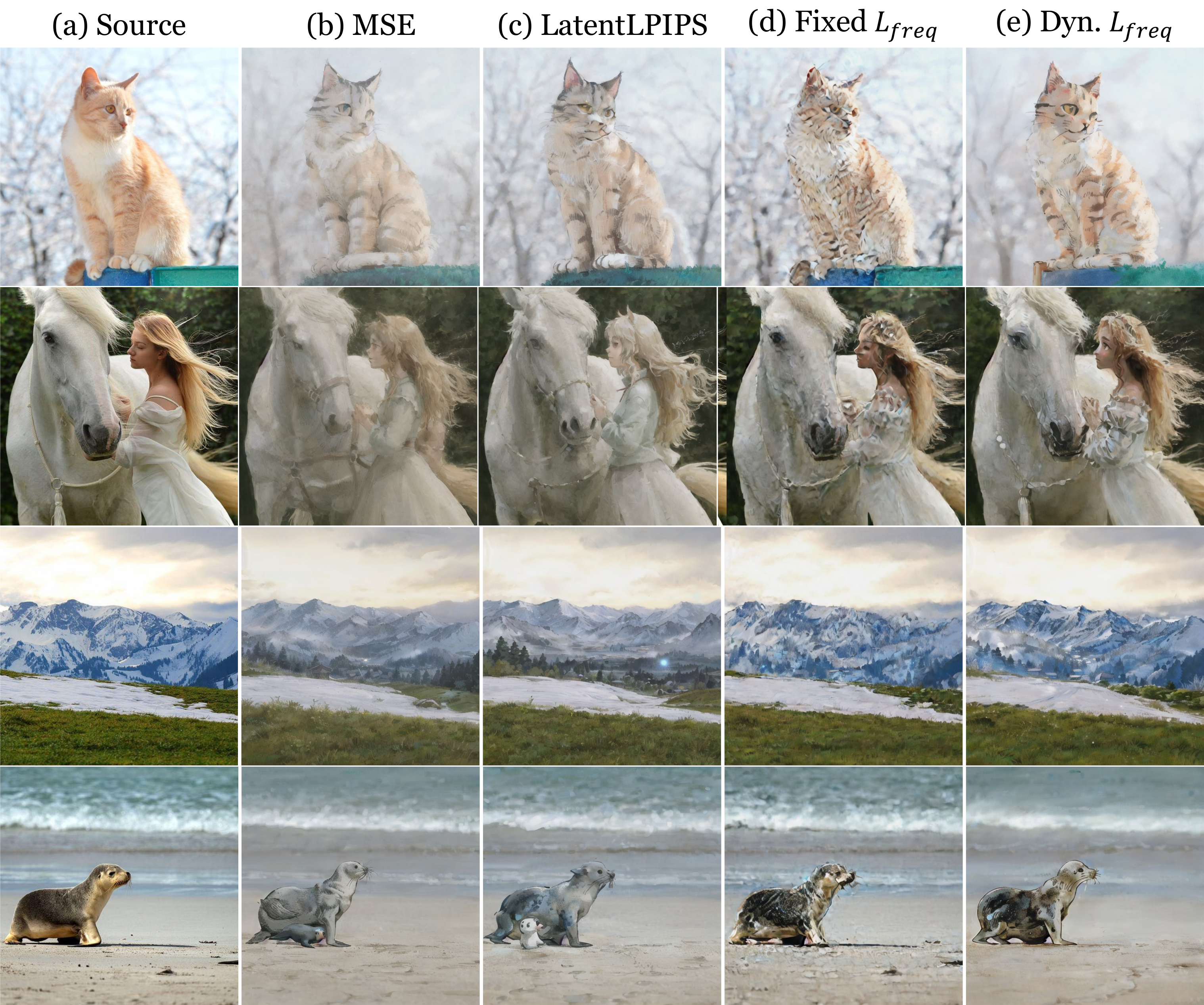}
    \vspace{-5mm}
    \caption{{\bf Comparison of identity loss variants.}} 
    \vspace{-6mm}
    \label{fig:ablationident}
\end{figure} 
\begin{figure*}[!t]
    \centering
    \vspace{-2mm}
    \includegraphics[width=0.98\linewidth]{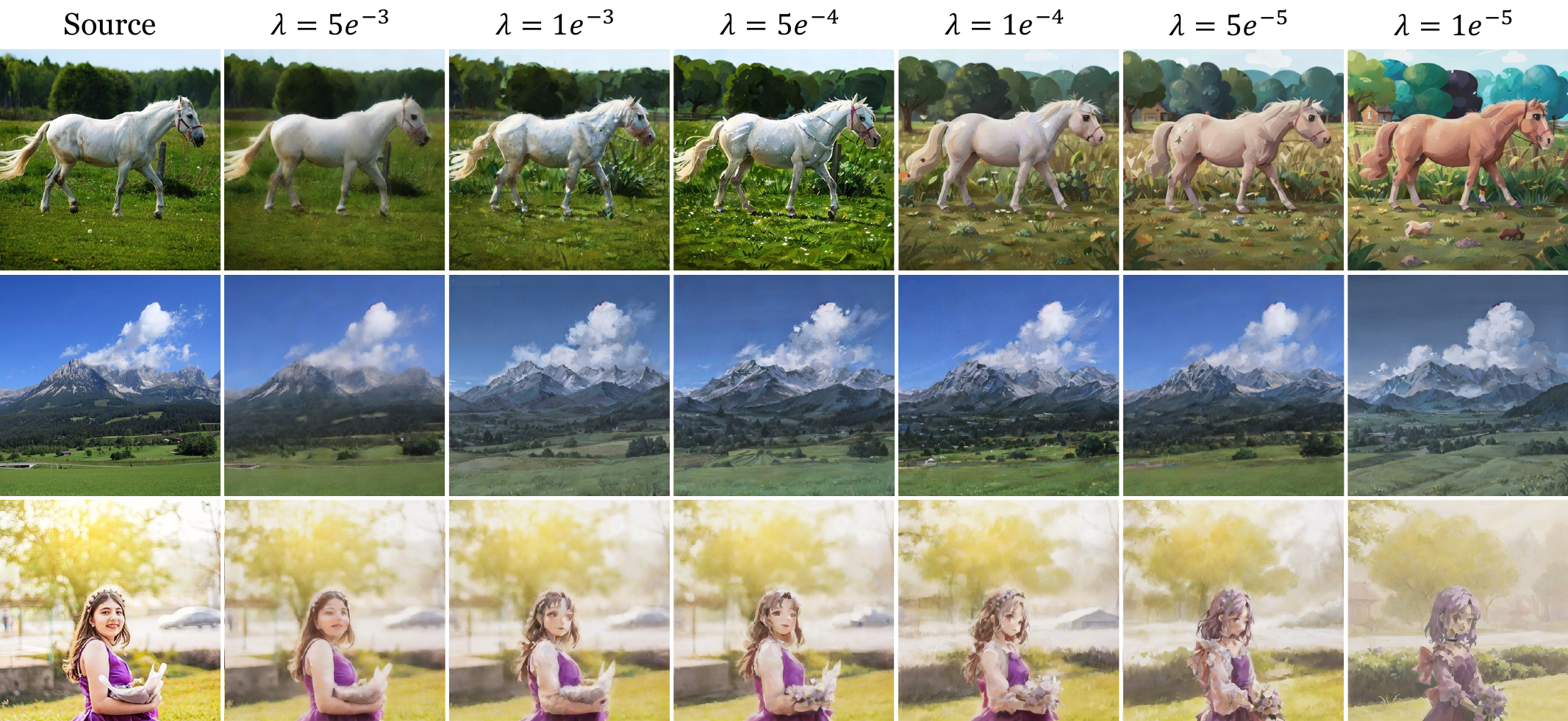}
    \vspace{-2mm}
    \caption{{\bf Effects of the loss weight $\lambda$.} The first row shows results in kids illustration style, and the last two rows show results in Ghibli style.} 
    \vspace{-1mm}
    \label{fig:lambda}    
\end{figure*}

The qualitative results, presented in Figure~\ref{fig:ablationident}, illustrate the limitations of these alternatives.
The MSE loss applied uniform regularization across all pixels, leading to blurriness and an inability to balance content fidelity with style adaptation (see Figure~\ref{fig:ablationident}(b)).
The LatentLPIPS loss, despite focusing on high-level feature alignment, struggles to maintain sufficient identity while incorporating style details. 
Adopting a fixed frequency cutoff results in oversharpened artifacts, underscoring the necessity of timestep-aware frequency regularization.
In contrast, our method successfully translates intricate high-frequency style textures, such as hairs details (see Figure~\ref{fig:ablationident}(e), Row $2$) and seal skin (Row $4$) while preserving low-frequency structure fidelity, like the mountain ridgeline (Row $3$).
Our progressive spectrum regularization strikes a balance between high-frequency style transfer fidelity and low-frequency content preservation.


\subsection{Effects of Loss Weight $\lambda$ Study}
The strength of the explicit identity regularization term is determined by the loss weight $\lambda$. 
As shown in Figure~\ref{fig:lambda}, increasing $\lambda$ enhances content fidelity, while reducing it allows for stronger stylization, demonstrating a clear trade-off between style and content. It provides a user-controllable knob for adjusting the stylization strength.

\section{More Qualitative Comparisons}
\label{sec:comparison}
\begin{figure*}[!h]
    \centering
    \includegraphics[width=\linewidth]{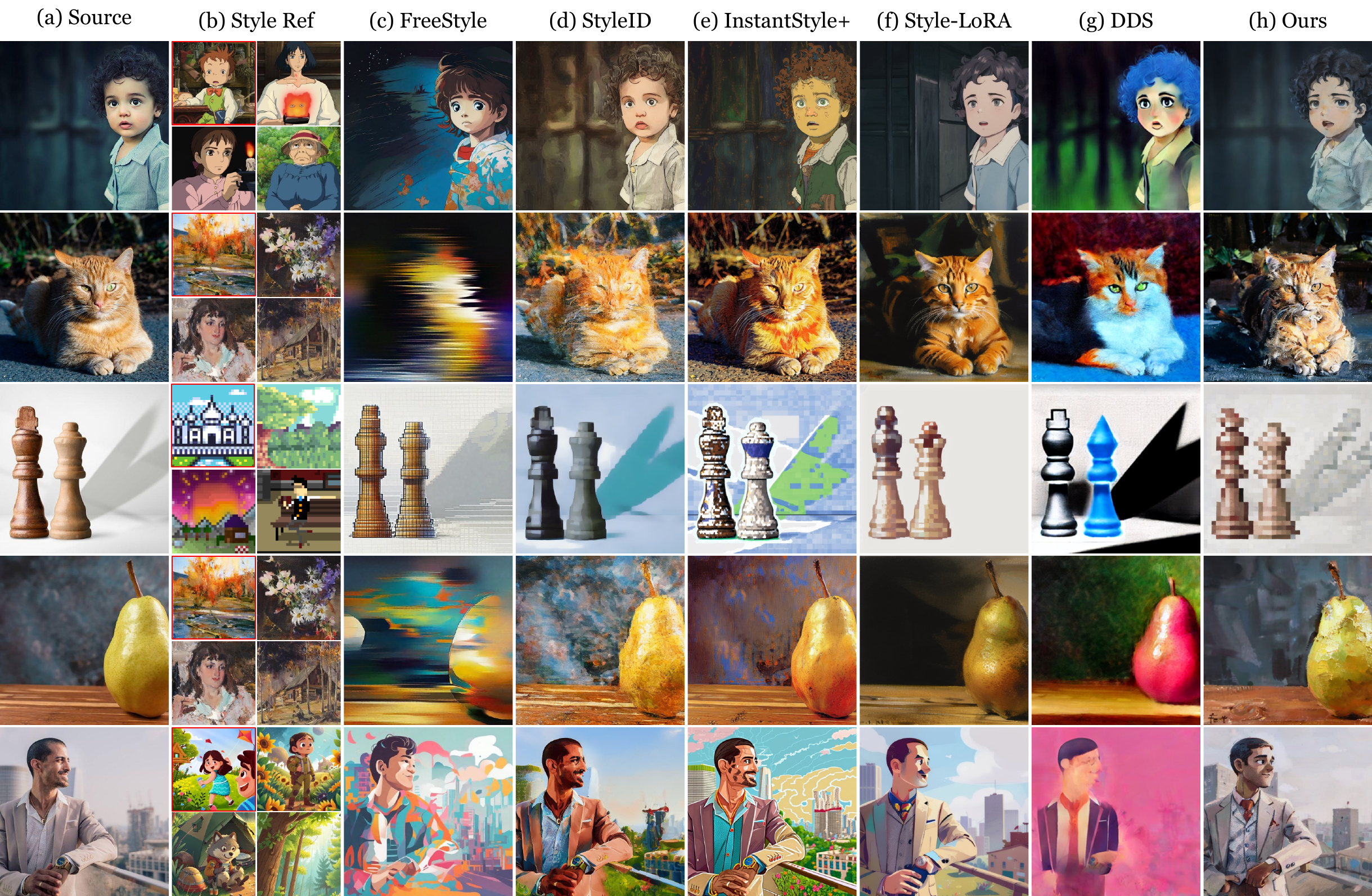}
    \caption{{\bf Additional qualitative comparison} between SMS (Ours) and five representative methods.}
    \vspace{-3.5mm}
    \label{fig:supp-comp}
\end{figure*}

%
We present additional qualitative comparisons with five state-of-the-art methods.
As shown in Figure~\ref{fig:supp-comp}, SMS achieves superior content preservation, maintaining structural integrity and ensuring a harmonious color balance, all while delivering comparable stylization results.

%

\section{Additional Results}
\label{sec:results}
\begin{figure*}[!h]
    \centering
    \includegraphics[width=0.85\linewidth]{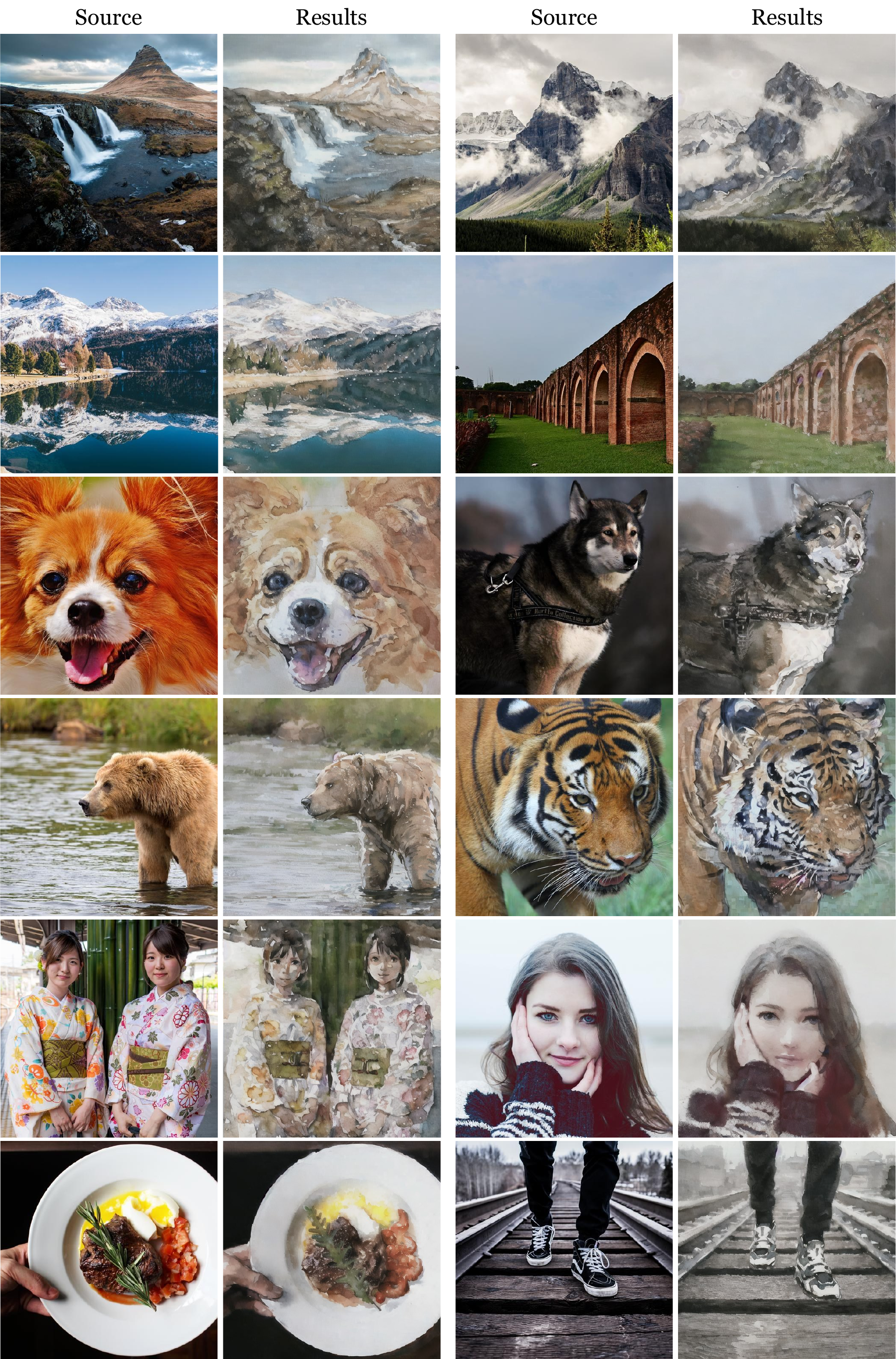}
    \caption{{\bf Watercolor style.} The results capture the fluid and translucent qualities typical of watercolor paintings, with gentle color gradients and soft edges.}
    \vspace{-3.5mm}
    \label{fig:watercolor}
\end{figure*}

\begin{figure*}[!h]
    \centering
    \includegraphics[width=0.85\linewidth]{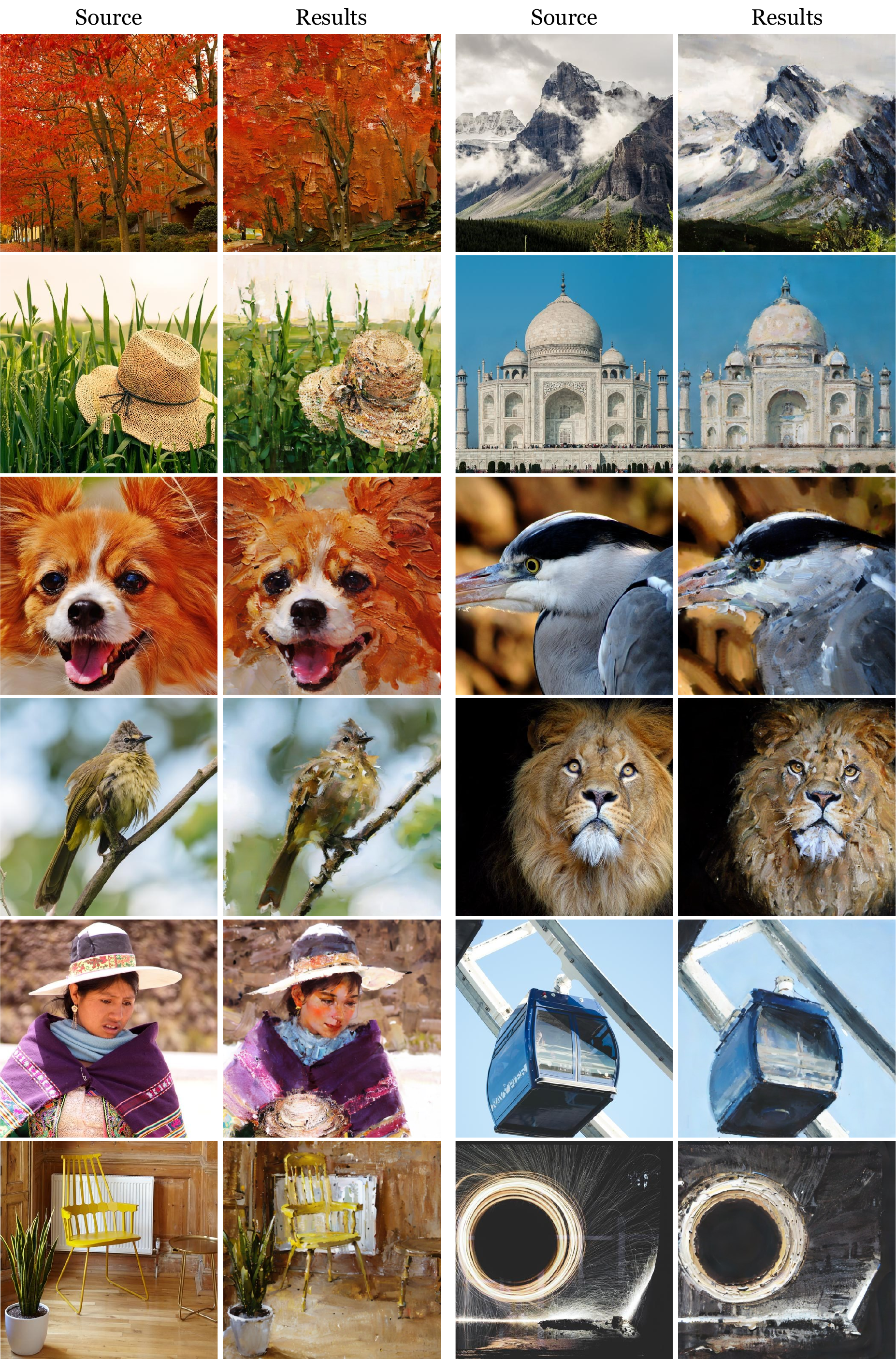}
    \caption{{\bf Oil painting style.} The results reflect the rich textures and bold brushstrokes associated with oil paintings, emphasizing depth and vibrancy.}
    \vspace{-3.5mm}
    \label{fig:oil}
\end{figure*}

\begin{figure*}[!h]
    \centering
    \includegraphics[width=\linewidth]{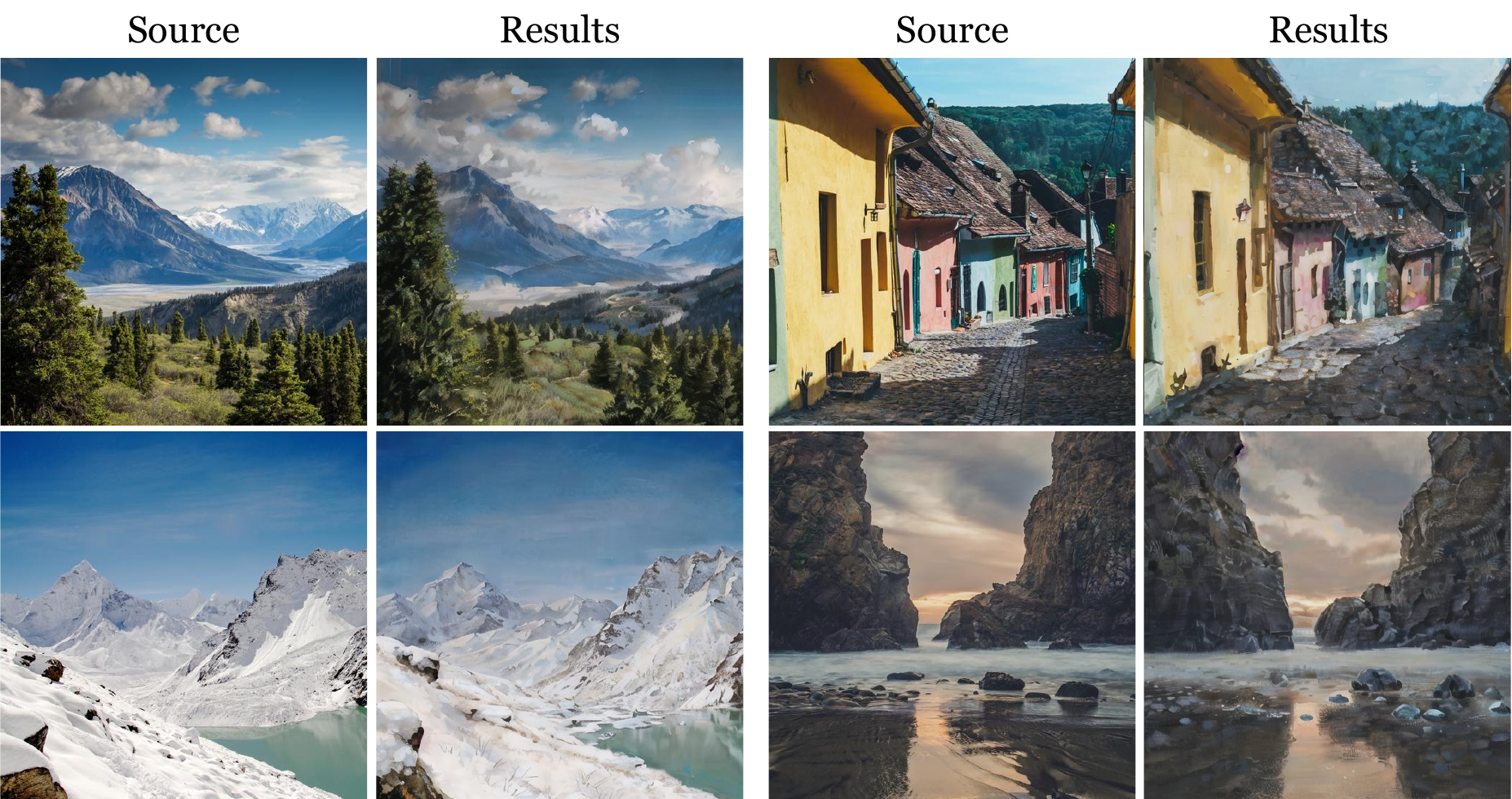}
    \caption{{\bf Ghibli style.} The results create a harmonious blend of realism and painterly artistry characteristic of Studio Ghibli, combining intricate pre-designed brush-like strokes in the scenes.}
    \vspace{-3.5mm}
    \label{fig:ghibli}
\end{figure*}

\begin{figure*}[!h]
    \centering
    \includegraphics[width=\linewidth]{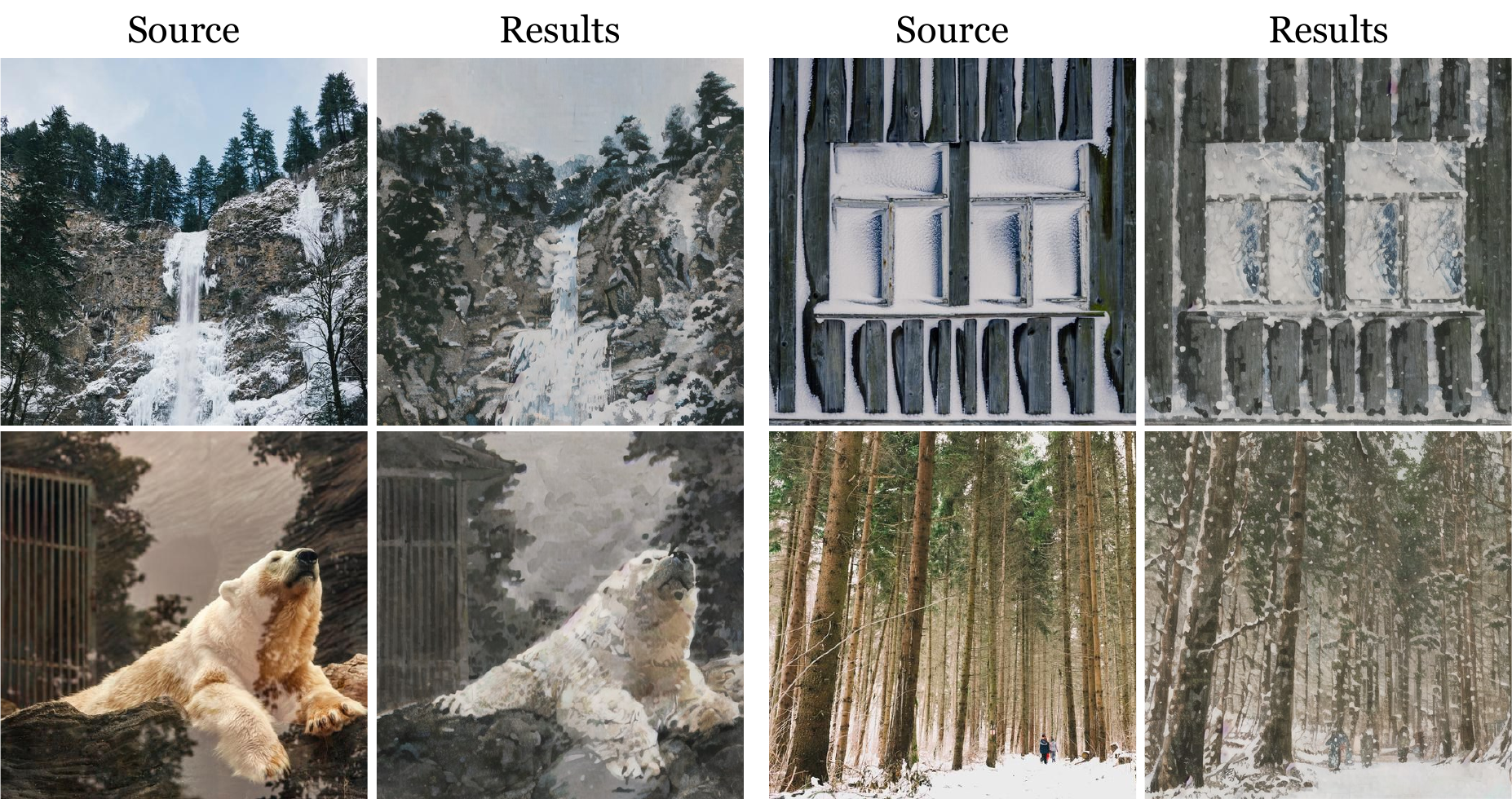}
    \caption{{\bf Ukiyo-e style.} The results reflect the essence of traditional Japanese ukiyo-e woodblock prints.}
    \vspace{-3.5mm}
    \label{fig:ukiyo}
\end{figure*}

\begin{figure*}[!h]
    \centering
    \includegraphics[width=\linewidth]{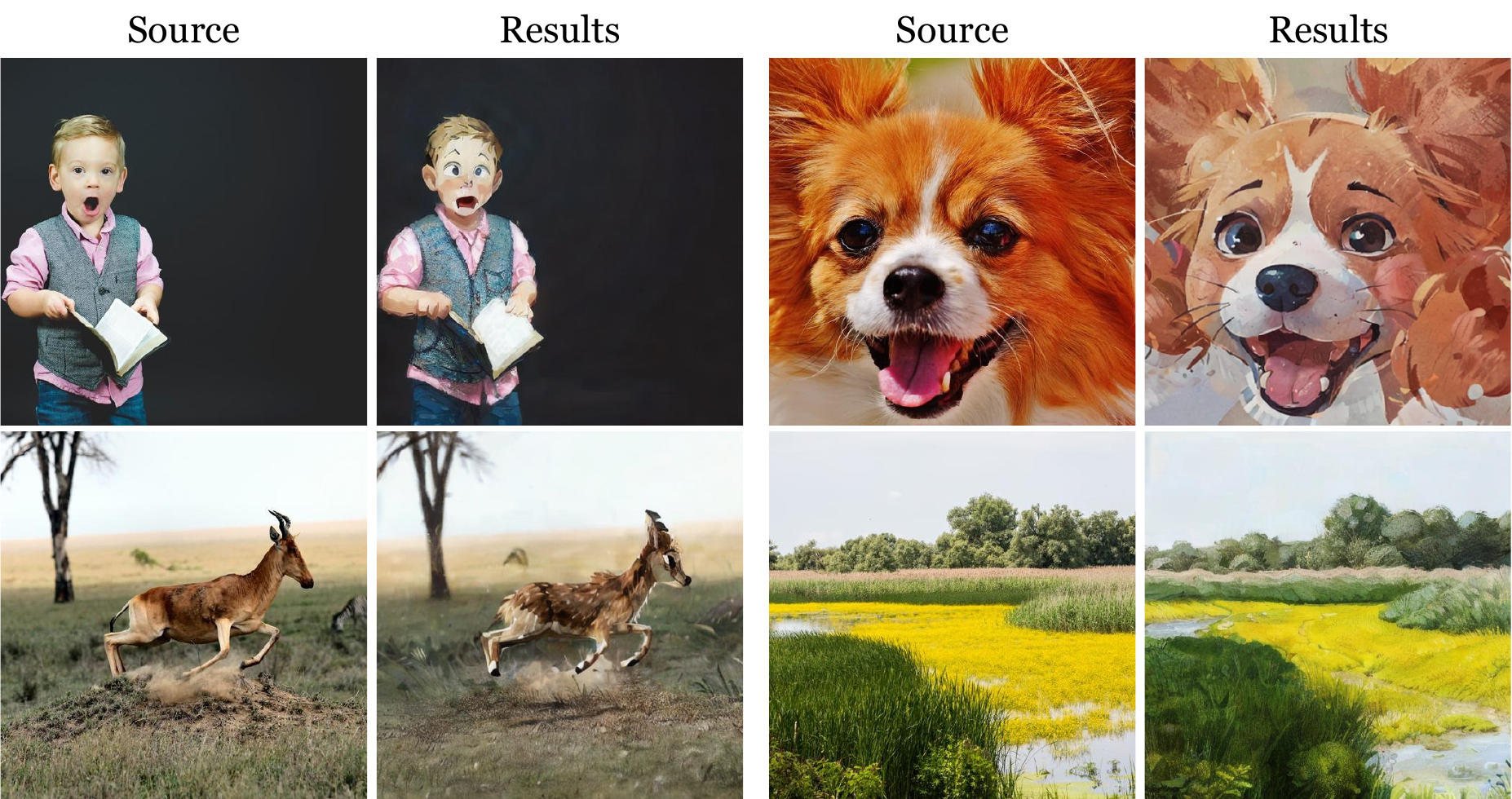}
    \caption{{\bf Kids illustration style.} The results have playful and vibrant qualities typical of children's illustrations, featuring simplified shapes and bold outlines.}
    \vspace{-3.5mm}
    \label{fig:kids}
\end{figure*}

\begin{figure*}[!h]
    \centering
    \includegraphics[width=\linewidth]{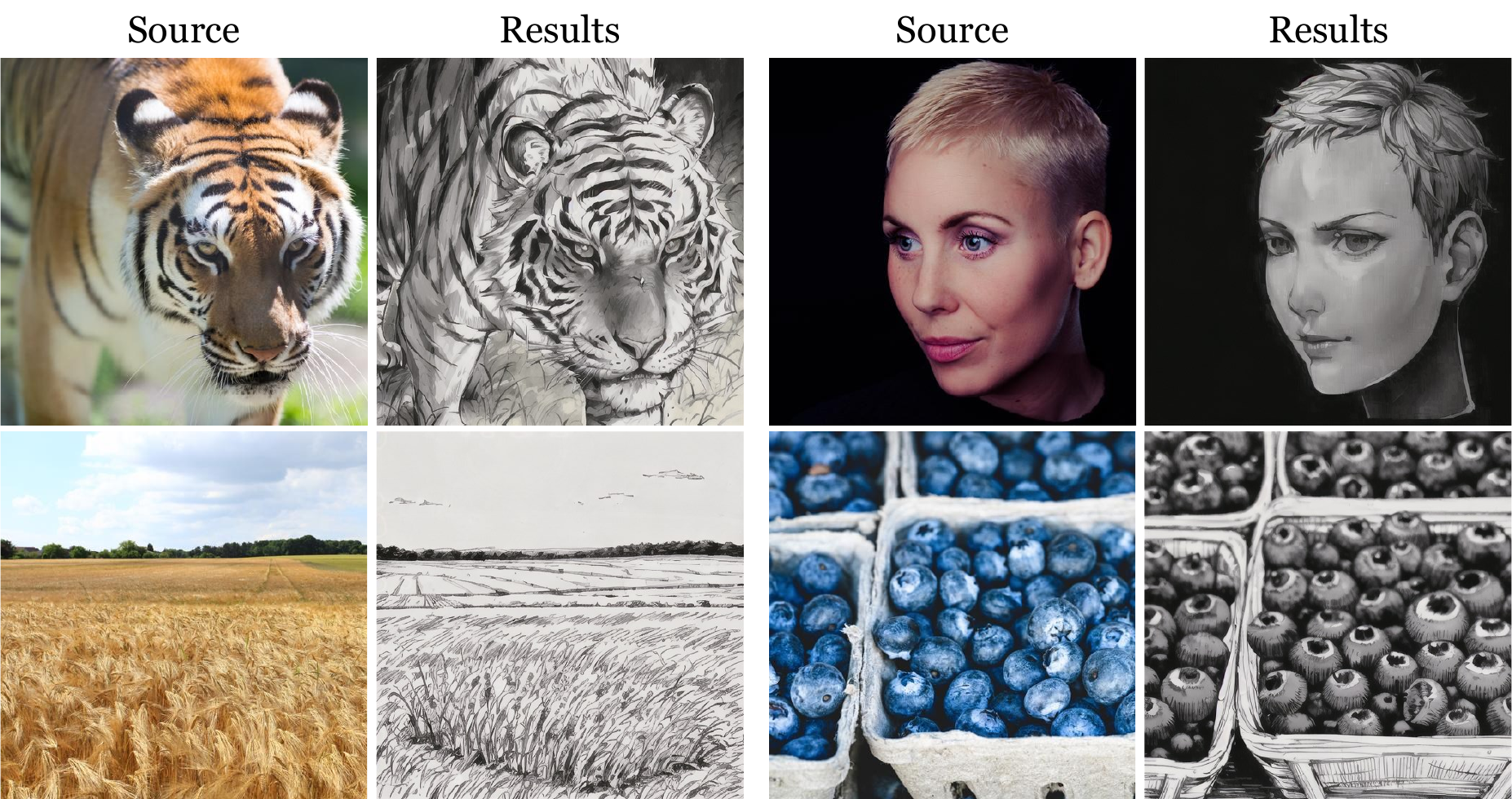}
    \caption{{\bf Sketch style.} The results resemble hand-drawn sketches, featuring monochromatic tones and emphasized contours.}
    \vspace{-3.5mm}
    \label{fig:sketch}
\end{figure*}

We provide additional examples of images generated by SMS on the DIV2K dataset~\cite{agustsson2017ntire} to showcase its superior high-quality balanced stylization ability across different styles. 
Figures~\ref{fig:watercolor}, \ref{fig:oil}, \ref{fig:ghibli}, \ref{fig:ukiyo}, \ref{fig:kids}, \ref{fig:sketch} displays stylizations in watercolor, oil painting, Ghibli, Ukiyo-e, kids illustration and sketch styles, respectively. 

\section{Limitations}
\label{sec:limitations}
\begin{figure}[!t]
    \centering
    \includegraphics[width=\linewidth]{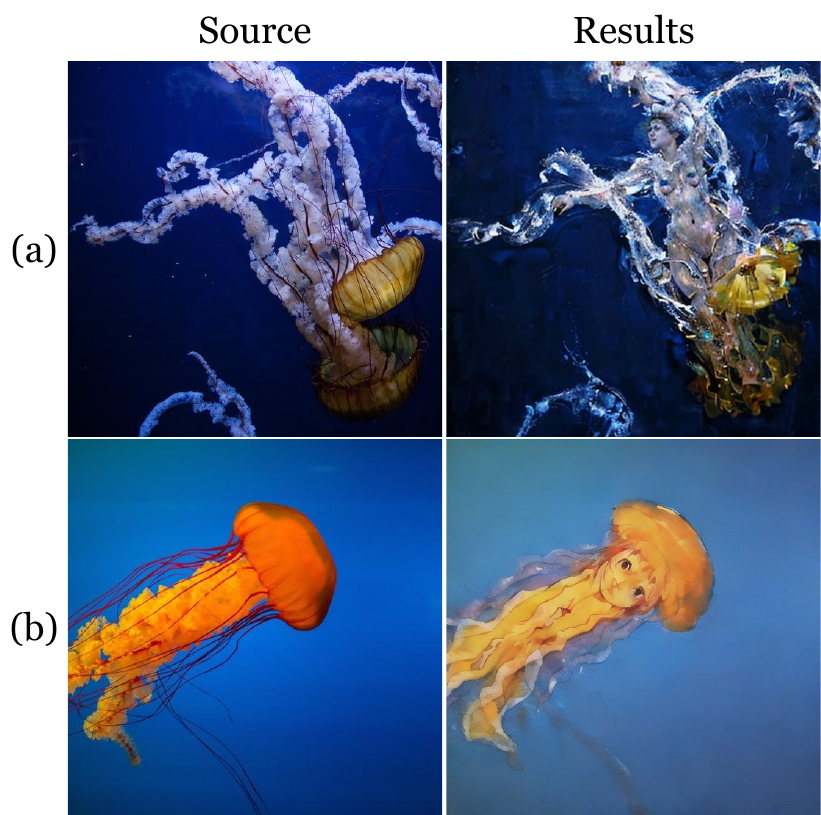}
    \vspace{-5.5mm}
    \caption{{\bf Certain failure cases.} (a) oil painting style with text prompt $y^\text{src}$: \textit{a group of jellyfish floating on top of a body of water.} (b) watercolor style with text prompt $y^\text{src}$: \textit{a jellyfish swimming in the ocean.}} 
    \vspace{-3.5mm}
    \label{fig:limitations}
\end{figure}

Despite the promising results, our method has certain limitations.
SMS relies on style-specific LoRAs, and if a LoRA lacks sufficient content diversity, especially for specific object categories, distortions may occur. 
For example, using an oil painting style LoRA that trained with few or no images of jellyfish can result in stylized outputs where jellyfish are inaccurately transformed into other objects, such as human figure (see Figure~\ref{fig:limitations}(a)).
This issue arises because the LoRA has not learned appropriate representations for those unseen or underrepresented content types.
Increasing the content preservation parameter $\lambda$ may mitigate this problem, albeit at the cost of reduced stylization strength.

\section{Broader Impact}
\label{sec:impact}
Our stylization framework has significant societal impacts. Positively, it can enhance creativity in graphic design, animation, and digital art, offering powerful tools for high-quality style transfer. It also holds promise for personalized education and immersive entertainment experiences.

However, we must be mindful of potential negative consequences. Biases present in training datasets can propagate through generative models, potentially amplifying societal inequities.
Furthermore, the ability to train a style-LoRA with limited artistic works and use SMS to transform other images into an artist's style raises concerns regarding intellectual property rights and copyright protection. Careful ethical considerations and adherence to copyright laws are crucial to mitigate these risks.


\end{document}